\documentclass{article}
\usepackage{iclr2021_workshop,times}

\usepackage{amsmath,amsfonts,bm}

\def\eqref#1{equation~\ref{#1}}

\def\1{\bm{1}}

\DeclareMathAlphabet{\mathsfit}{\encodingdefault}{\sfdefault}{m}{sl}
\SetMathAlphabet{\mathsfit}{bold}{\encodingdefault}{\sfdefault}{bx}{n}

\usepackage{hyperref}
\usepackage{url}
\usepackage{booktabs}
\usepackage[textwidth=3cm]{todonotes}

\title{Geometry encoding \\for numerical simulations}

\author{Amir Maleki$^1$, Jan Heyse$^2$, Rishikesh Ranade$^1$, Haiyang He$^1$, Priya Kasimbeg$^1$ \& Jay Pathak$^1$\\
{$^1$Ansys Inc, $^2$Stanford University}\\
\texttt{\{amir.maleki, rishikesh.ranade, haiyang.he \& jay.pathak\}@ansys.com}\\
\texttt{heyse@stanford.edu}
}

\iclrfinalcopy 
\begin{document}

\maketitle

\begin{abstract}
We present a notion of geometry encoding suitable for machine learning-based numerical simulation. In particular, we delineate how this notion of encoding is different than other encoding algorithms commonly used in other disciplines such as computer vision and computer graphics. We also present a model comprised of multiple neural networks including a processor, a compressor and an evaluator. These parts each satisfy a particular requirement of our encoding. We compare our encoding model with the analogous models in the literature.
\end{abstract}

\section{Introduction}

Applications of machine learning for accelerating and replacing numerical simulations have received significant traction over the past few years. Various physics-based \citep{raissi2018hidden, wu2018physics, raissi2019physics, rao2020physics, gao2020phygeonet, qian2020lift}, data-driven \citep{morton2018deep, thuerey2020deep, pfaff2020learning} as well as hybrid algorithms \citep{xue2020amortized, ranade2020discretizationnet, kochkov2021machine}  are proposed for this purpose. The existing studies are primarily tested on simple geometries and academic benchmark problems; e.g. flow over a cylinder or sphere. While these problems play a fundamental role in developing intuition and shed light on how machine learning can assist numerical simulations, they are significantly simpler than problems we encounter in real-life applications, and fail to generalize to different geometries even with same underlying physical problem. Here, we aim to move towards achieving machine learning-based algorithms that are capable of solving partial differential equation over complex domains, which generalize across various complex geometries and assemblies. More specifically, we introduce the notion \emph{geometry encoding} for \emph{numerical simulations}: a methodology to encode complex geometric objects using neural networks to create a compressed spatial representation. This notion of geometry encoding can be used for i) solving partial differential equation using machine learning, ii) discretization and meshing the geometry, and iii) \emph{a priori} computational resource prediction (e.g. for cloud computing).

Geometrical objects are ubiquitous in a wide range of disciplines including computer vision and perception, and computer graphics. Hence, various learning and non-learning based algorithms for geometry representation and geometry encoding are present in the literature. However, geometry encoding for numerical simulations demands a specific set of requirements. As an example, in computer vision, one may only care about the boundaries of the object present in geometry, and therefore, encoding is more focused on resolving the boundary of objects. However, in the case a numerical simulation, a geometry encoding algorithm has to be accurate not only at the object boundary, but also the entire domain. In \S \ref{sec:reqs} we lay down the specific feature of geometry encoding for numerical simulations. 

\subsection{Requirements of geometry encoding for numerical simulations}\label{sec:reqs}

An encoded geometry for the purpose of numerical simulations should have the following features: 
\subsubsection{Global accuracy of encoding} As explained earlier, for the purpose of numerical simulations, geometry encoding not only concerns the objects boundaries, but also their respective distances, as well as their distance from the domain boundaries. This differentiates our notion of geometry encoding from those commonly used in computer vision and computer graphics \citep{eslami2018neuralrep, mescheder2019occupancy, park2019deepsdf, gropp2020implicit}, where object boundaries receive significantly more attention. As an example, consider the flow over an sphere; one of the most iconic example of solving Navier-Stokes equations. Geometry encoding needs to accurately resolve not only the boundaries of the sphere, but also the boundaries of the entire domain (inlets, outlets or walls), where fluid boundary conditions and source terms may be specified. 

\subsubsection{Compressed encoding}
A popular approach for numerically solving partial differential equations using machine learning is to reduce the problem dimensionality and to process in an encoded space, where the operations are faster, and the training is simpler \citep{morton2018deep, carlberg2019recovering, he2020unsupervised,  ranade2020discretizationnet}. Compressing the geometry from a sparse high-dimensional representation to a lower-dimensional dense representation will lead to faster learning and more robust generalization. In addition to numerical simulations, a condensed representation of geometry can be useful for tasks such as predicting computational resources needed for a particular simulation (e.g. for cloud computing). For these reasons, the encoded geometry should be represented in a lower dimensional manifold. Note that compression should take place without losing the accuracy of representation.

\subsubsection{Continuity and differentiability of encoding}
Many industrial, environmental or biological processes can be accurately modeled with partial differential equations (PDE). In fact, that is why the ability to solve these equations have been pursued for many decades. In particular, accurate and rapid PDE solvers can be used for design purposes, where several design parameters can change, often times leading to a computationally infeasible design space. Gradient-based optimization algorithms can be utilized to efficiently search the design space \citep{morton2018deep, remelli2020meshsdf}. Therefore, an ideal geometry encoding should be continuous and differentiable with respect to the parameters of geometry (e.g. coordinate variables). This feature enables us to further use machine learning to identify optimal designs. 

\subsubsection{Variable encoding/decoding resolution}
Geometrical objects in the context of numerical simulation may have different levels of complexity. Instead of encoding all input geometries to a fixed-size encoding, an ideal geometry encoder for numerical simulation may use a variable size encoding depending on the complexity of the geometry. In addition, on the decoding side, depending on the applications, the geometry encoder should be capable of reconstructing the geometry with a variable resolution.

\section{Signed Distance Field as a geometry encoding}

Over the past few years, several studies have focused on identifying memory-efficient, yet expressive, means to represent 2D and 3D geometries. Continuous implicit representation has particularly received significant attention \citep{mescheder2019occupancy, park2019deepsdf, duan2020curriculumsdf, chibane2020unsignedsdf, sitzmann2020metasdf}. As pointed out earlier in \S \ref{sec:reqs}, a continuous representation can also be used for optimization purposes \citep{remelli2020meshsdf}. 

In this paper, we argue the signed distance field (SDF) can be used as a geometry encoder for numerical simulation with the requirements specified in \S \ref{sec:reqs}. Consider a geometry that contains one or multiple objects. The signed distance of any point within the geometry is defined as the distance between that point and the boundary of the closest object.

Below, we describe our model of a geometry encoder for numerical simulations. We assume the geometry files are provided as binary 2D images. We discuss the limitation of this assumption in our concluding remarks \S \ref{sec:conc}. Our geometry encoding model is comprised of three parts: i) processor; ii) compressor and iii) evaluator:
\subsection{Processor}
The processor consists of a U-net \citep{ronneberger2015u} configuration. We used strided convolutional layers to down-sample the input image resolution, while expanding the feature channels. The up-sampling is performed using transposed convolutional layers (see Fig.~\ref{fig:unet}). The input to the processor is the binary images, and the output is the SDF. The role of the processor is to \emph{accurately} perform the transition from binary pixel data to SDF, satisfying one of the requirements of the geometry encoder. 

An important aspect of the processor configuration is the skip connections that connect earlier layers before the latent code to later layers of the network after the latent code. In our experience, we were not able to train a processor with reasonable accuracy without the skip connections. The presence of the skip connection means that the processor is not able to provide a compressed encoding, because the intermediate values before the latent code (the skip connections) are necessary for decoding the encoded geometry. Therefore, the role of compressed encoding is assigned to the second part: the compressor. 

\subsection{compressor}
The second part of the geometry encoding is the compressor network which receives the output of the processor as input, and returns a true compressed encoding. The compressor has a structure similar to the processor, except with no skip connection (i.e. a convolutional autoencoder). In our experiments presented in \S \ref{sec:exp}, the compressor compresses the input by a factor of $8$, with insignificant loss of accuracy. We believe more efficient compressors can be trained if the hyper-parameters and model configurations are optimized. We did not pursue this route.

\begin{figure}[t]
\centering
\begin{tabular}{cc}
    {binary image \hspace{0.5cm} ground truth \hspace{1.2cm} MetaSDF \hspace{1cm} processor \hspace{.7cm} compressor}\\
	\includegraphics[trim=0cm 1cm 1cm 1cm, clip=true, scale=0.23]{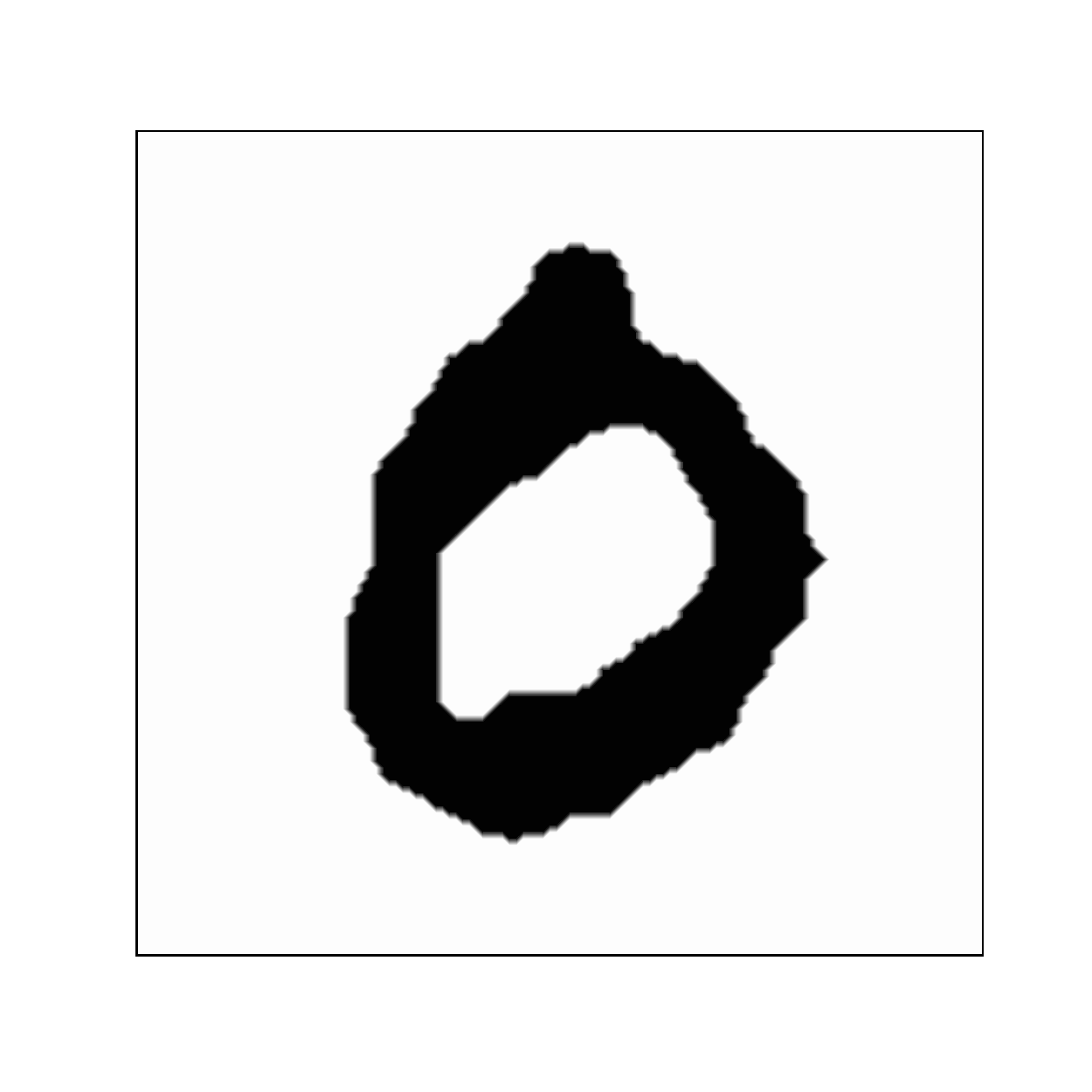}
	\hspace{-0.5cm}
	\includegraphics[trim=0cm 1cm 1cm 1cm, clip=true, scale=0.23]{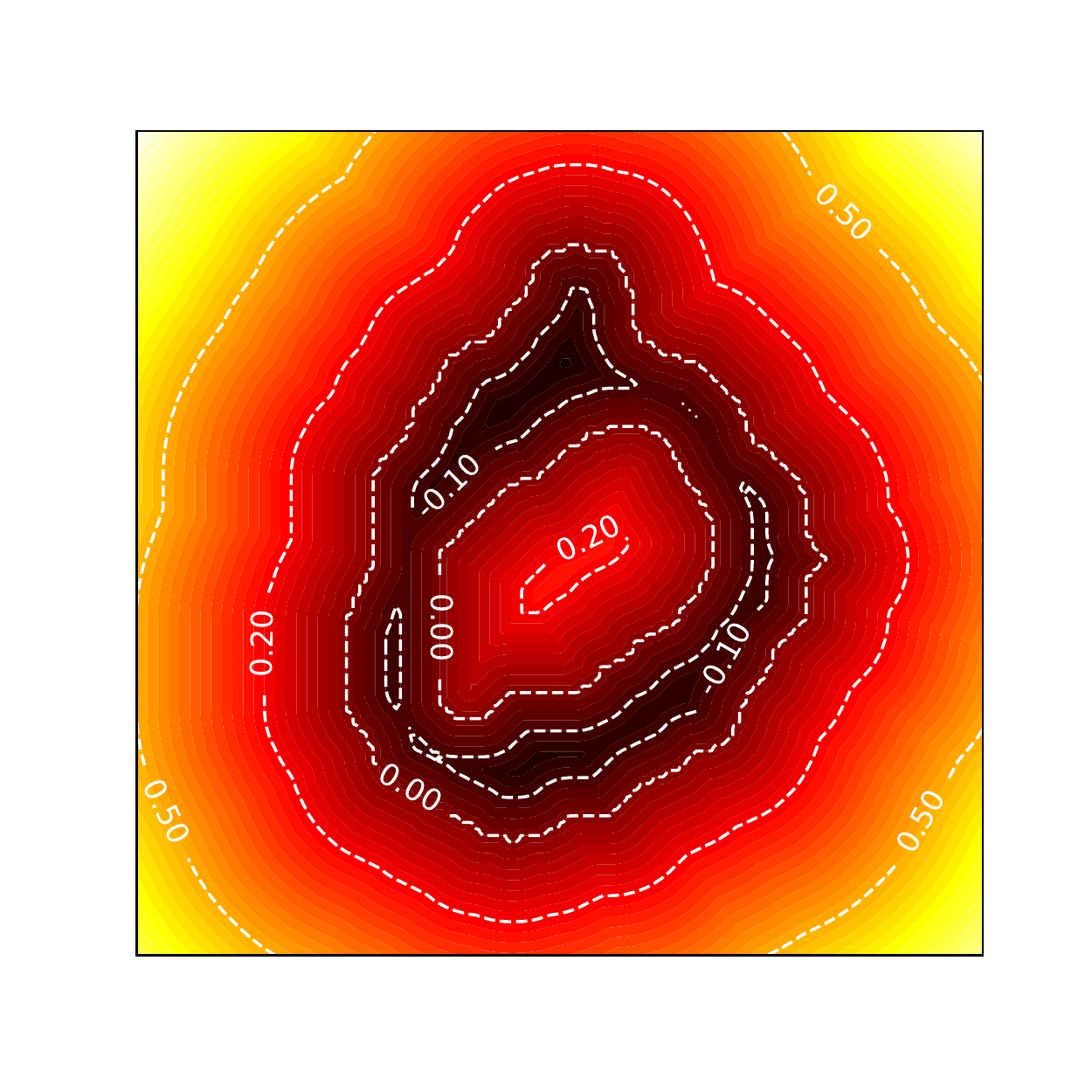}
	\hspace{0.25cm}
	\includegraphics[trim=0cm 1cm 1cm 1cm, clip=true, scale=0.23]{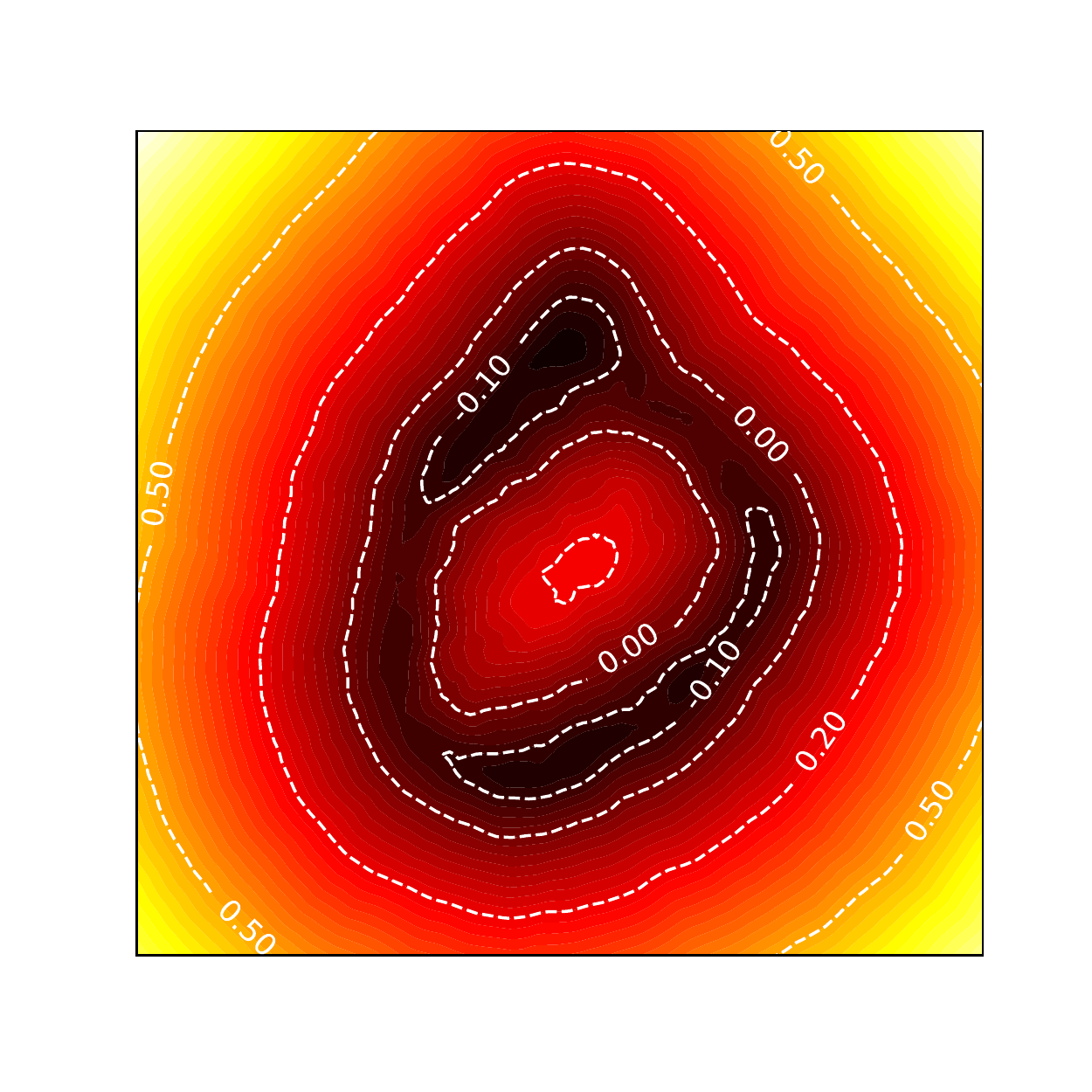}
	\hspace{-0.4cm}
	\includegraphics[trim=0cm 1cm 1cm 1cm, clip=true, scale=0.23]{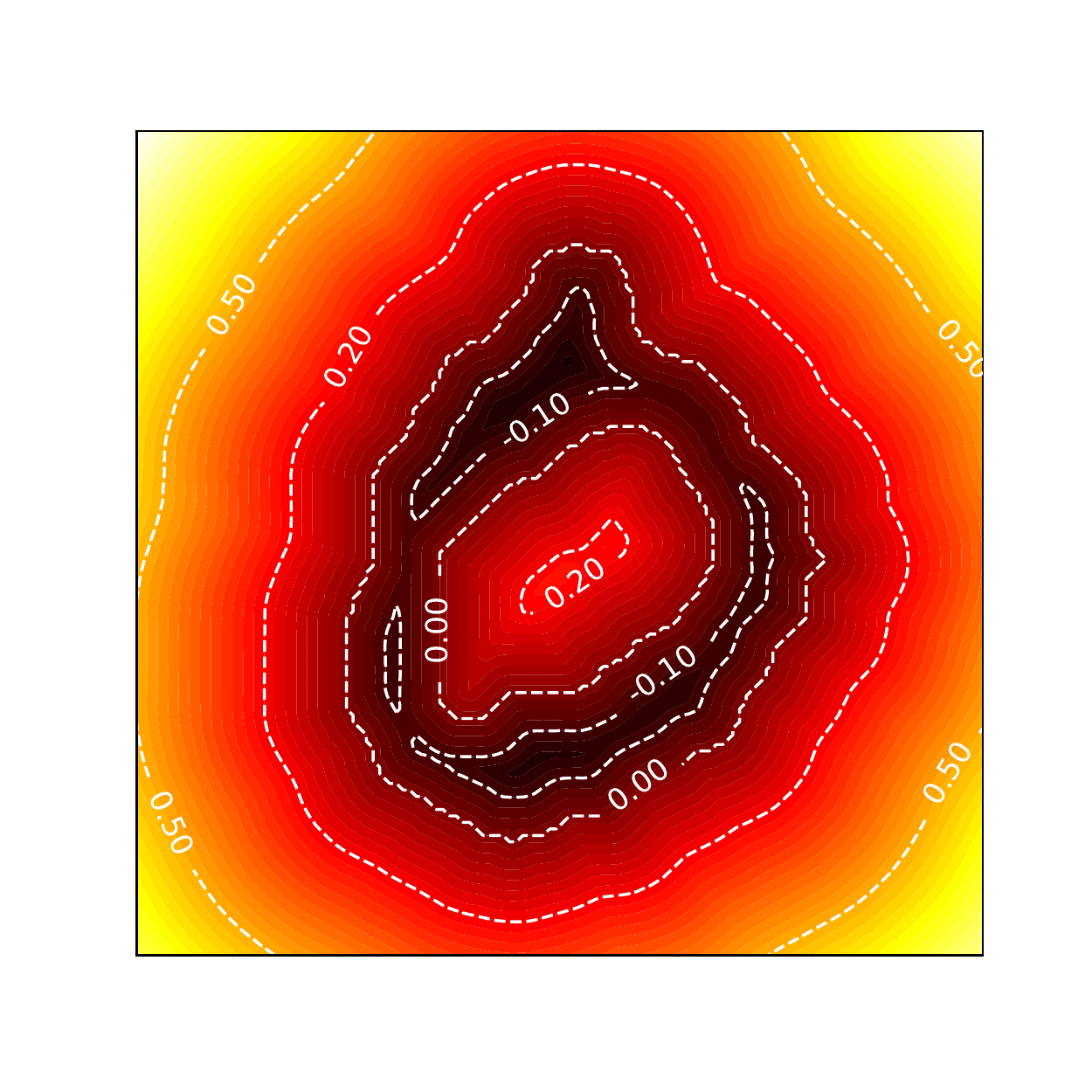}
	\hspace{-0.4cm}
	\includegraphics[trim=0cm 1cm 1cm 1cm, clip=true, scale=0.23]{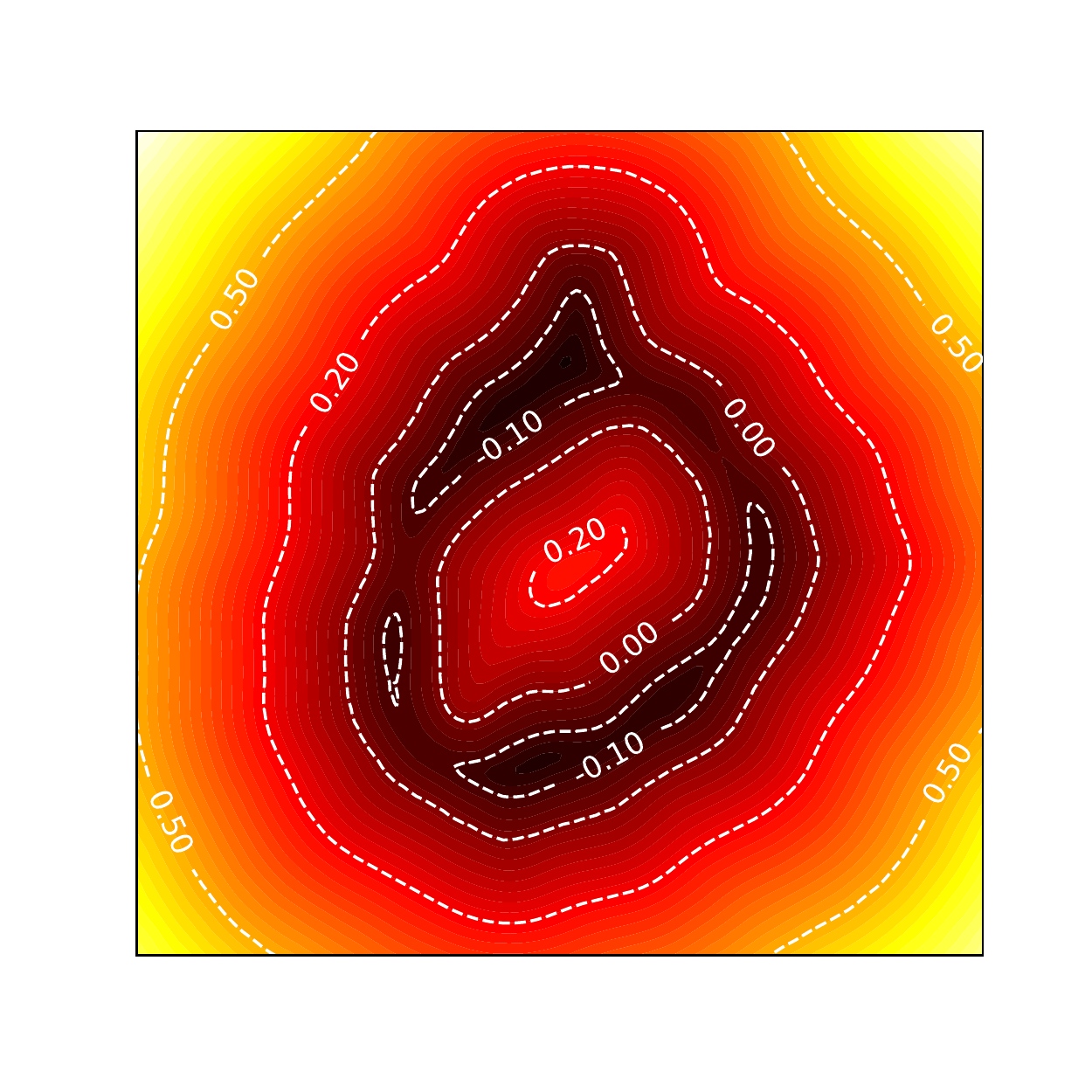}\\
	
	\includegraphics[trim=0cm 1cm 1cm 1cm, clip=true, scale=0.23]{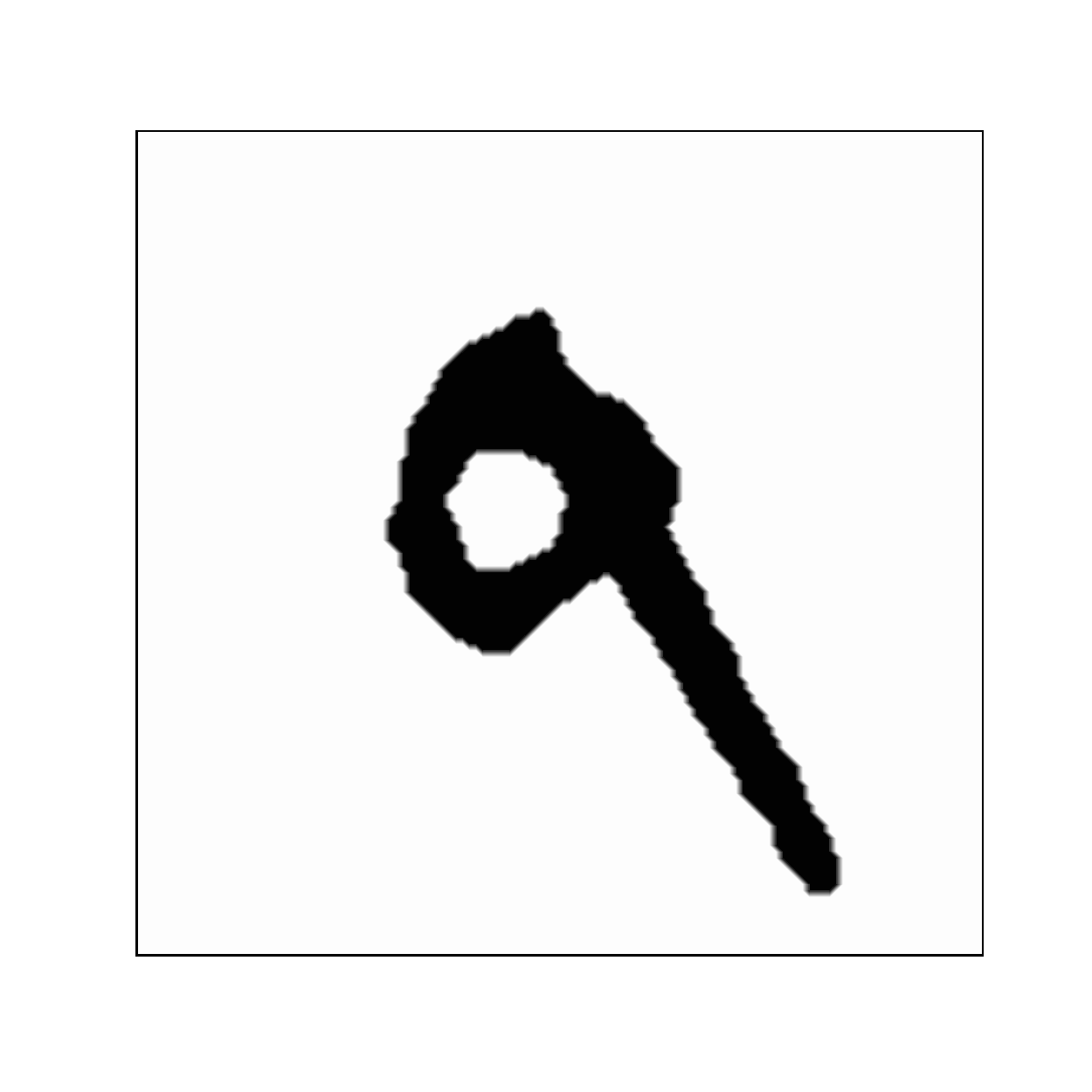}
	\hspace{-0.5cm}
	\includegraphics[trim=0cm 1cm 1cm 1cm, clip=true, scale=0.23]{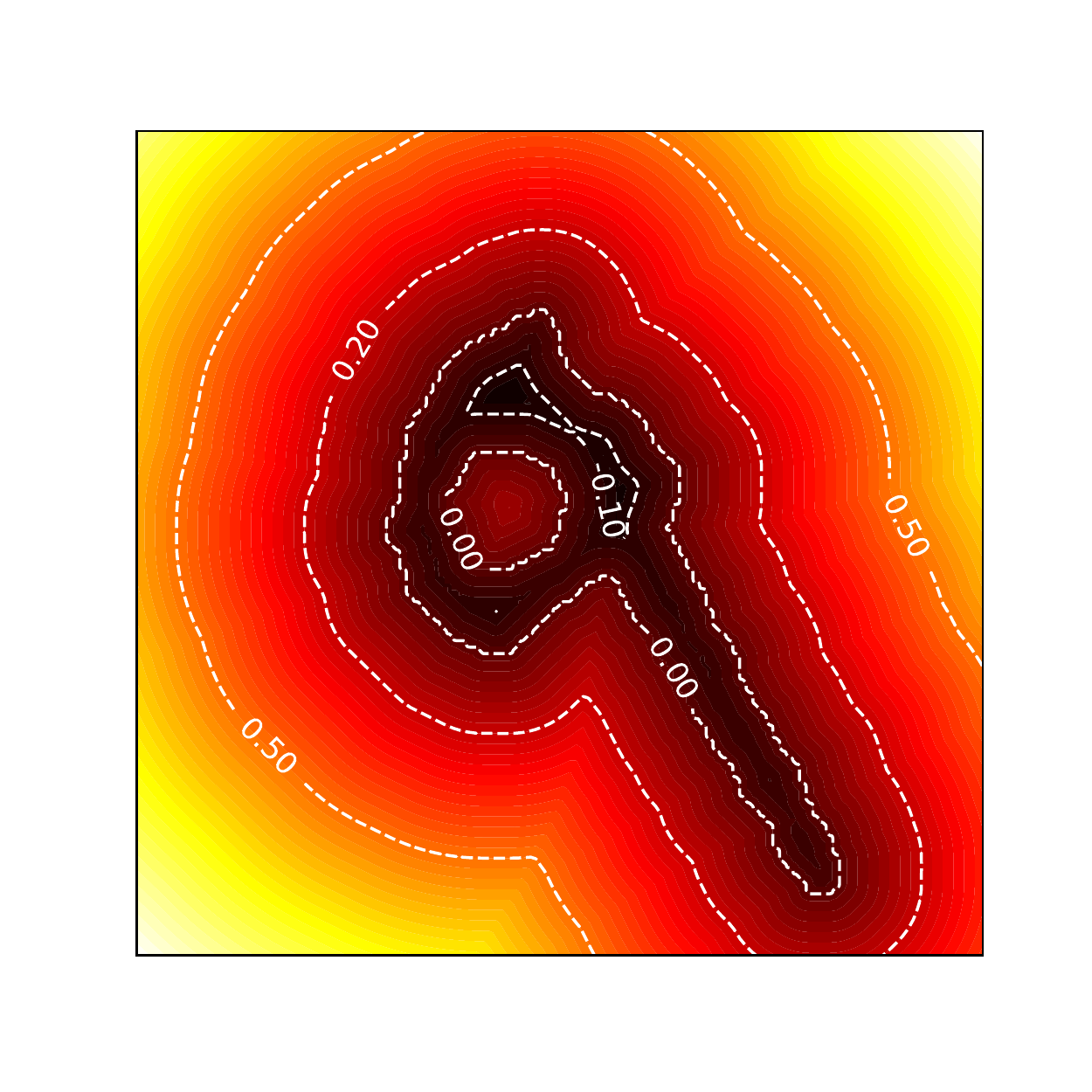}
	\hspace{0.25cm}
	\includegraphics[trim=0cm 1cm 1cm 1cm, clip=true, scale=0.23]{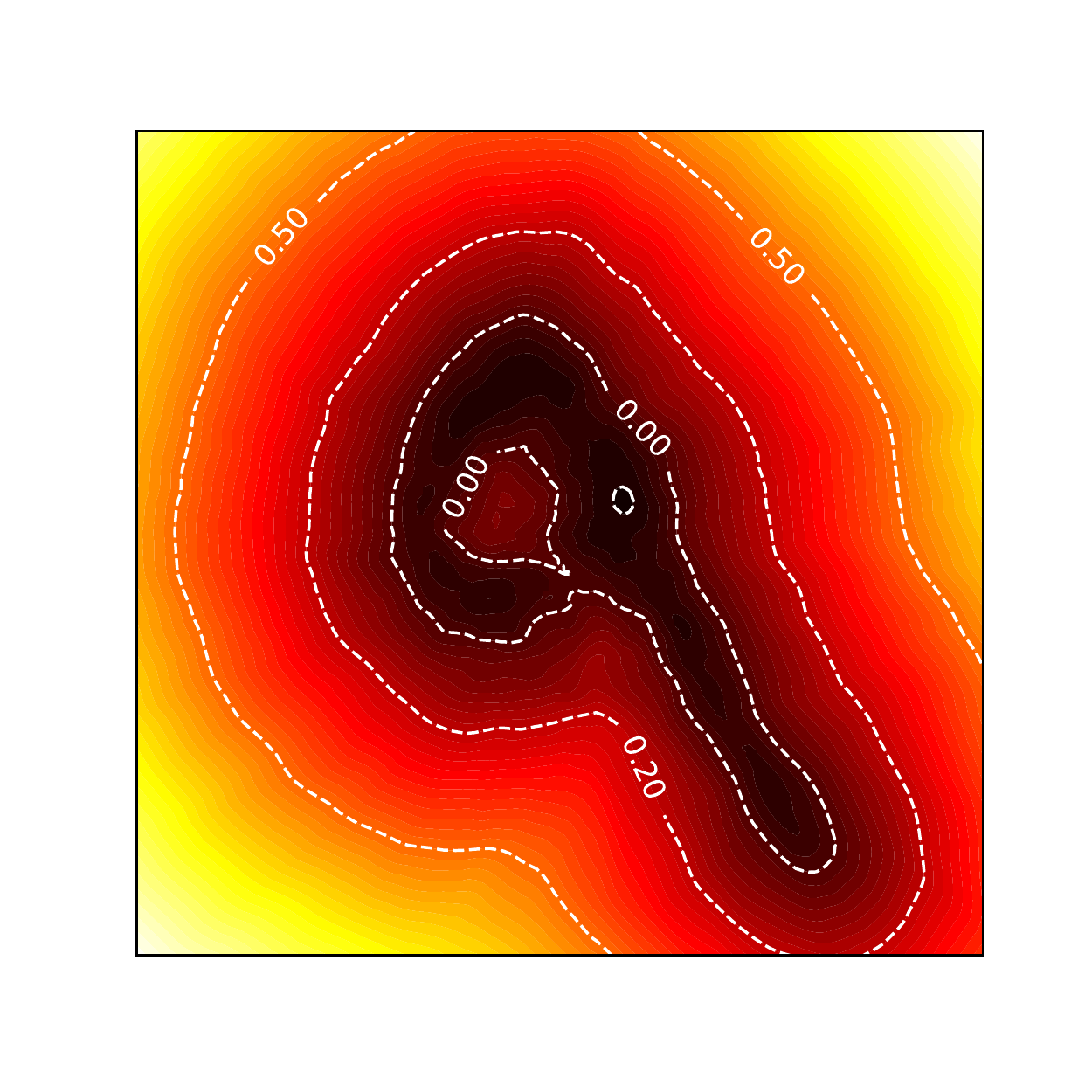}
	\hspace{-0.4cm}
	\includegraphics[trim=0cm 1cm 1cm 1cm, clip=true, scale=0.23]{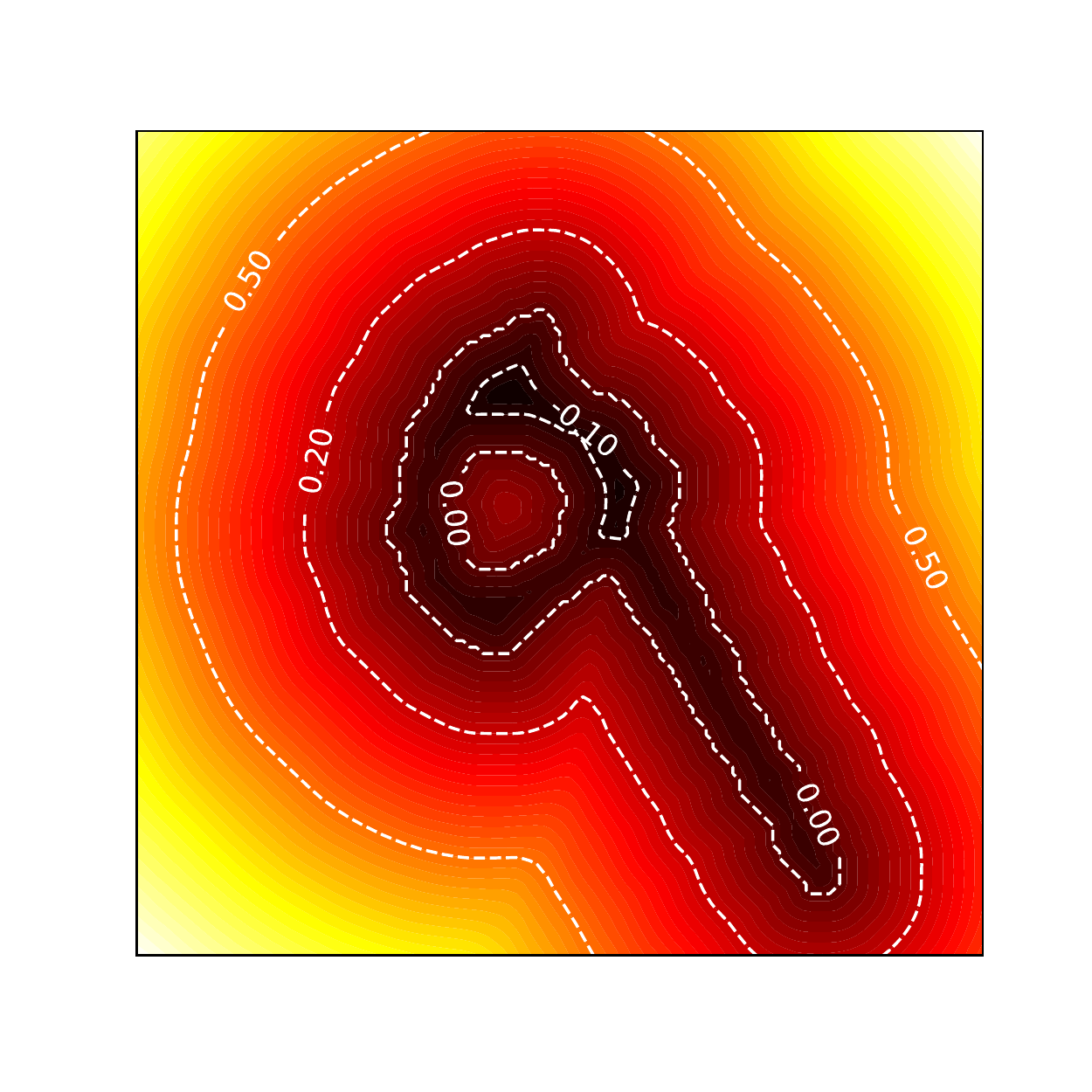}
	\hspace{-0.4cm}
	\includegraphics[trim=0cm 1cm 1cm 1cm, clip=true, scale=0.23]{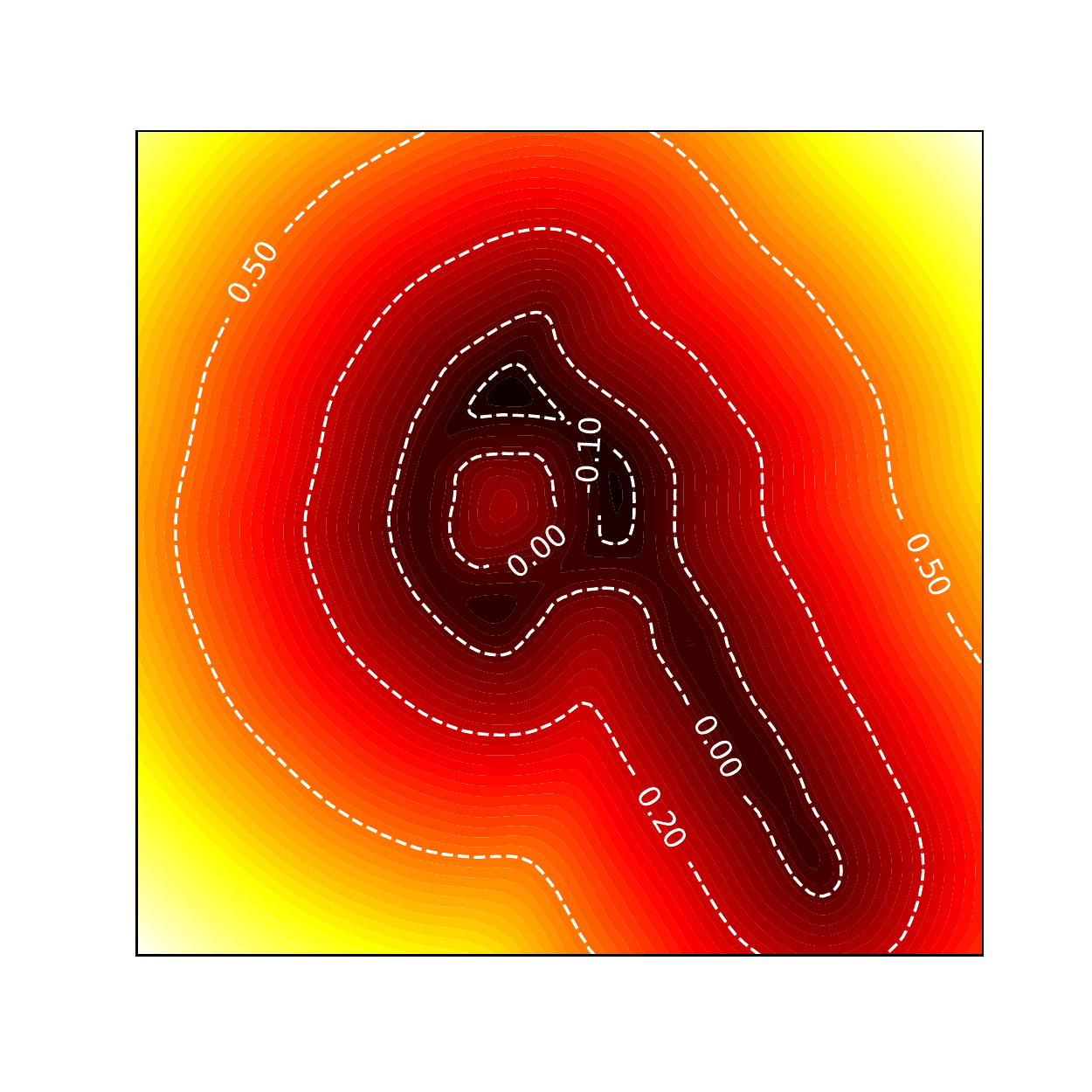}
	\\
	 {\hspace{-0.4cm} MetaSDF error \hspace{1.1cm} processor error \hspace{1.6cm} compressor error }
	 \\
	\includegraphics[trim=0cm 1cm 0cm 0cm, clip=true, scale=0.35]{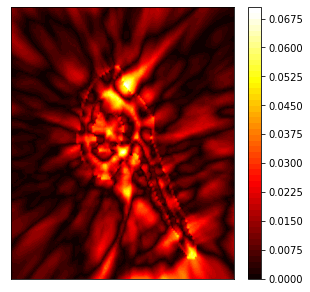}
	\includegraphics[trim=0cm 1cm 0cm 0cm, clip=true, scale=0.35]{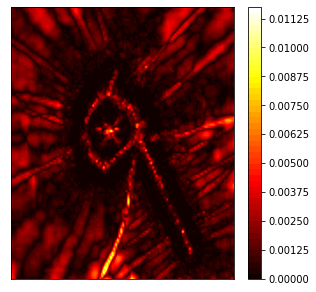}
	\includegraphics[trim=0cm 1cm 0cm 0cm, clip=true, scale=0.35]{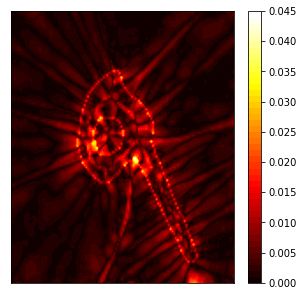}
\end{tabular}
\caption{Comparison of our work (processor and compressor models) against MetaSDF \citep{sitzmann2020metasdf} on MNIST dataset. White dashed lines show the iso-line contours.  The compressor generates an 8x smaller encoding compared to the processor. The bottom row shows absolute mean error.}
\label{fig:comparison}
\end{figure}

\subsection{evaluator}

In order for the geometry encoder to be used in an optimization scheme, we need to specify an evaluator. The role of the evaluator is to interpolate the SDF solution at any given $(x, y)$ coordinate points. The evaluator should be differentiable. 

For practical purposes, the evaluator implementation should support \emph{automatic differentiation}. Otherwise, it cannot be used in a deep learning based optimization procedure. As an example, a pure python implementation does not support automatic differentiation. In this study, we implemented a bilinear interpolation with no learning parameters within the PyTorch \citep{paszke2019pytorch} framework. This is the simplest design for an evaluator, and can be improved by adding learning parameters and/or using more complex interpolating schemes.

\section{Comparison}\label{sec:exp}

The accompanying code can be found at \href{https://github.com/ansysresearch/geometry_encoding}{https://github.com/ansysresearch/geometry-encoding}. The details of data generation and the training procedure is outlined in \ref{sec:training}. In particular, we emphasize that our synthesized training dataset is comprised of \emph{only} circles and polygons. However, the trained model learns to produce accurate prediction on more complex geometries.

To evaluate how well our geometry encoding works, we compare our results with those of MetaSDF paper \citep{sitzmann2020metasdf} on the MNIST dataset. The MetaSDF study is one of the most recent publications regarding learning SDF, and generates results on par with other contenders. Fig.~\ref{fig:comparison} shows two examples of the MNIST dataset in the validation set. The MetaSDF is directly trained on the MNIST dataset, and the results are generated via meta-learning (i.e. it requires a few steps of gradient descent at inference time). Our model, however, has only been trained on our synthesized primitive dataset (Fig.~\ref{fig:datagen}), which includes no shapes similar to MNIST dataset. Despite such a difference, Fig.\ref{fig:comparison} shows that our model performs more accurately. In particular, we emphasize how our model preserves the levelsets of SDF. We have also prepared a more quantitative comparison  over a validation dataset containing 500 examples. The aggregate results are shown in Table \ref{table:comparison}. 

For more results on other datasets, refer to \S \ref{sec:exotic}.

\begin{table}[t]
\caption{Comparison of MetaSDF with the processor and compressor networks on MNIST dataset over 500 samples of a validation set. The compressor generates an 8x smaller encoding compared to the processor. }
\label{table:comparison}
\begin{center}
\begin{tabular}{@{}lcccc@{}}
\toprule
  & MetaSDF                       & Processor            & Compressor  \\ \midrule
$L_1$      & $4.3 \times 10^{-3}$  & $4.3 \times 10^{-4}$ & $1.4 \times 10^{-3}$   \\
$L_2$      & $3.6 \times 10^{-5}$  & $5.4 \times 10^{-7}$ & $4.7 \times 10^{-6}$    \\
$L_\infty$ & $7.8 \times 10^{-2}$  & $3.9 \times 10^{-2}$ & $5.0 \times 10^{-2}$ 
\end{tabular}
\end{center}
\end{table}

\section{Conclusion}\label{sec:conc}
In this paper, we introduced the idea of geometry encoding for the purpose of numerical simulation with four specific features: i) global accuracy; ii) compressed encoding; iii) differentiability with respect to geometry parameters and iv) variable-size encoding. We presented a simple neural network structure comprised of three parts: processor, compressor and evaluator which satisfy the first three features. We also compared our results with that of MetaSDF \citep{sitzmann2020metasdf}.

This paper was intended to be an introduction to the idea of geometry encoding for numerical simulations. We aim to extend this idea in multiple different directions. Most immediately, we intend to establish a similar encoder for 3D voxelized data. Other 2D and 3D representations such as point clouds and meshes are more prevalent in simulations. We intend to extend this analysis to these geometry representations. In particular, we believe graph neural networks are suitable for working with mesh data. Moving towards working with meshed input data will also allow us to satisfy the fourth feature: variable-size encoding. 

\newpage
\bibliography{references}
\bibliographystyle{iclr2021_workshop}

\newpage

\appendix
\section{Appendix}
\subsection{data generation and training}\label{sec:training}

\begin{figure}[h]
\begin{center}
\includegraphics[scale=0.45, angle=0, trim={0cm 0cm 0cm 0cm}]{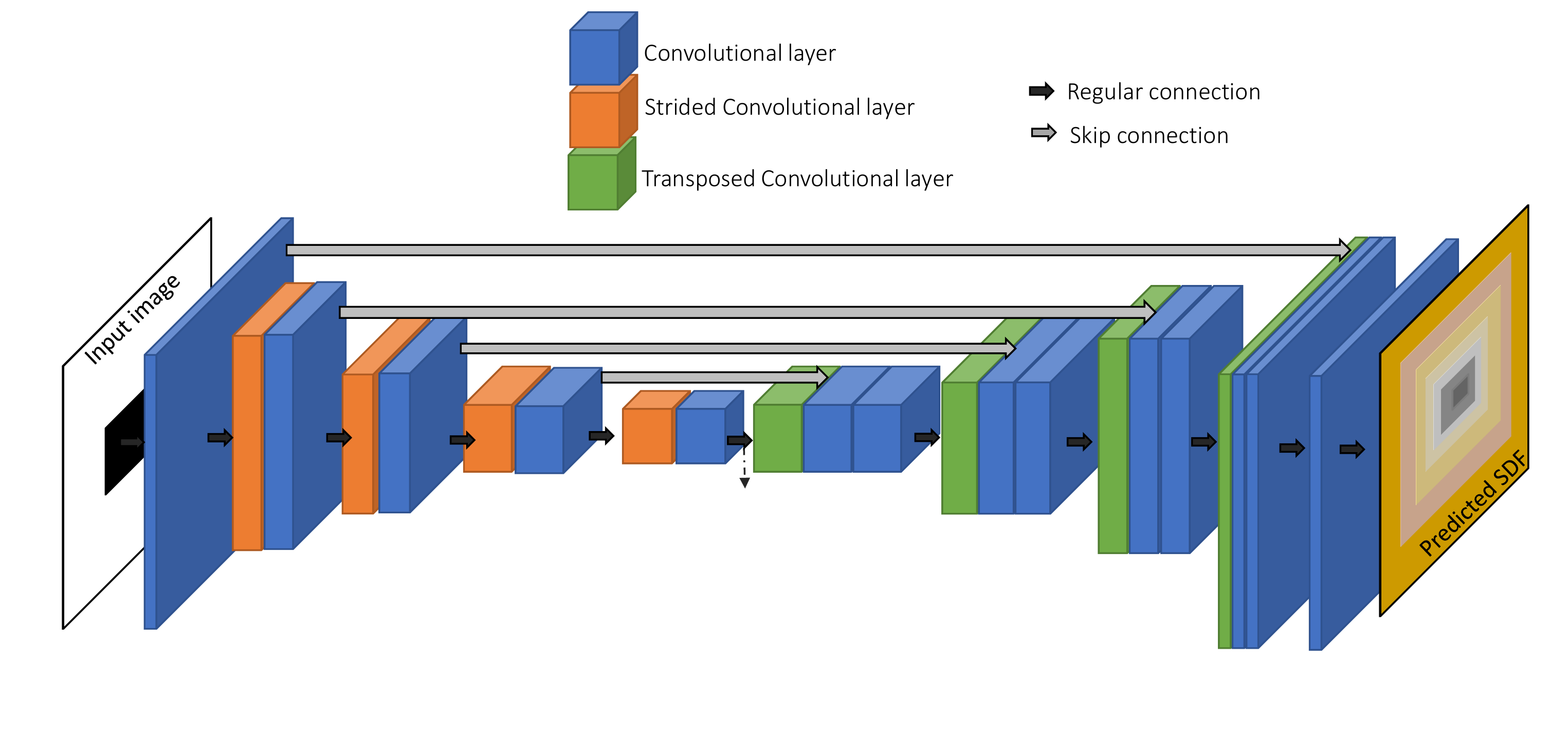}
\caption{The configuration of processor and compressor networks. The skip connections are absent in the compressor network. \label{fig:unet}}

\end{center}
\end{figure}

\begin{figure}[t]
\centering
\begin{tabular}{cc}
    {\hspace{-9.5cm} binary image \hspace{0.5cm} SDF }\\
	\includegraphics[trim=0cm 1cm 1cm 1cm, clip=true, scale=0.20]{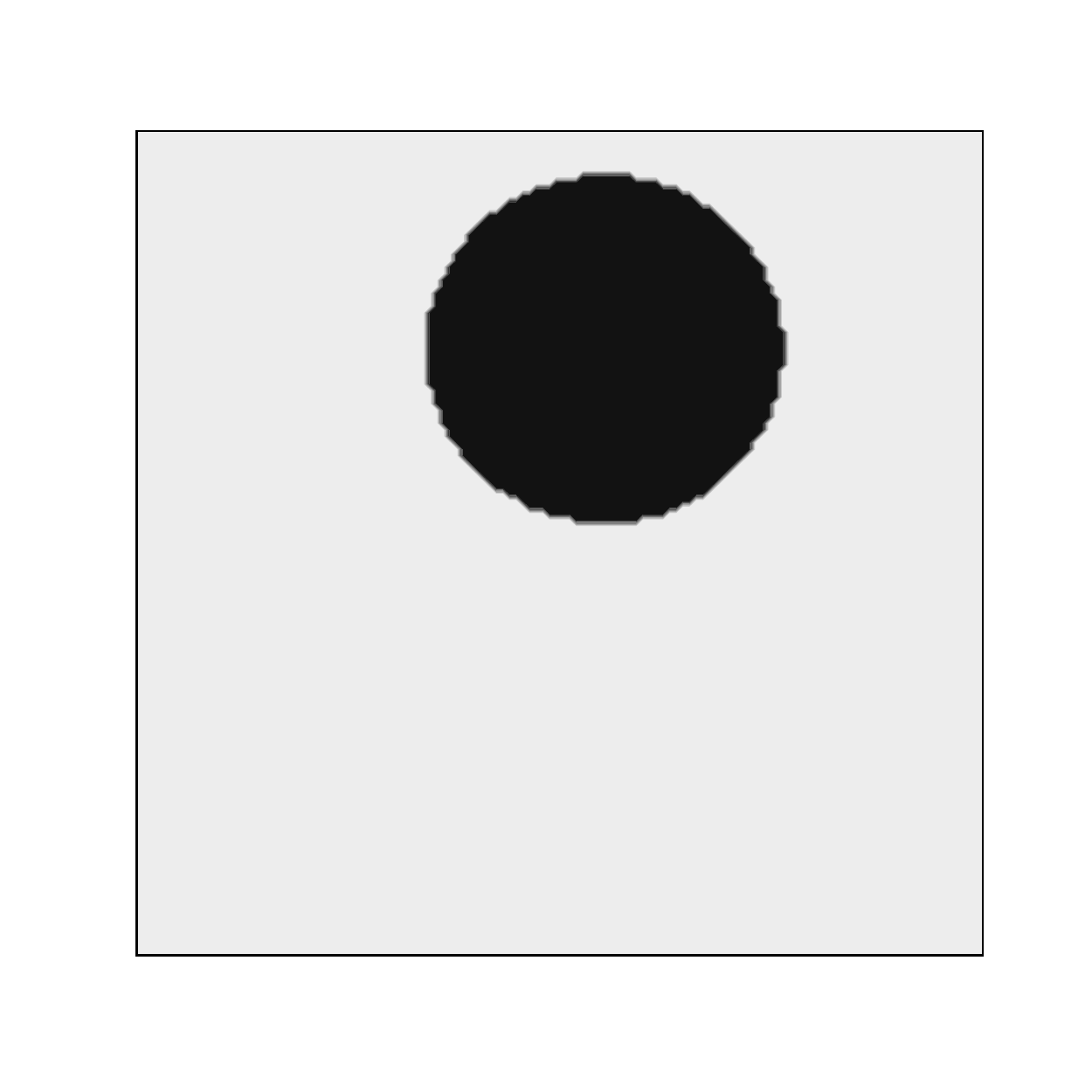}
	\hspace{-0.5cm}
	\includegraphics[trim=0cm 1cm 1cm 1cm, clip=true, scale=0.20]{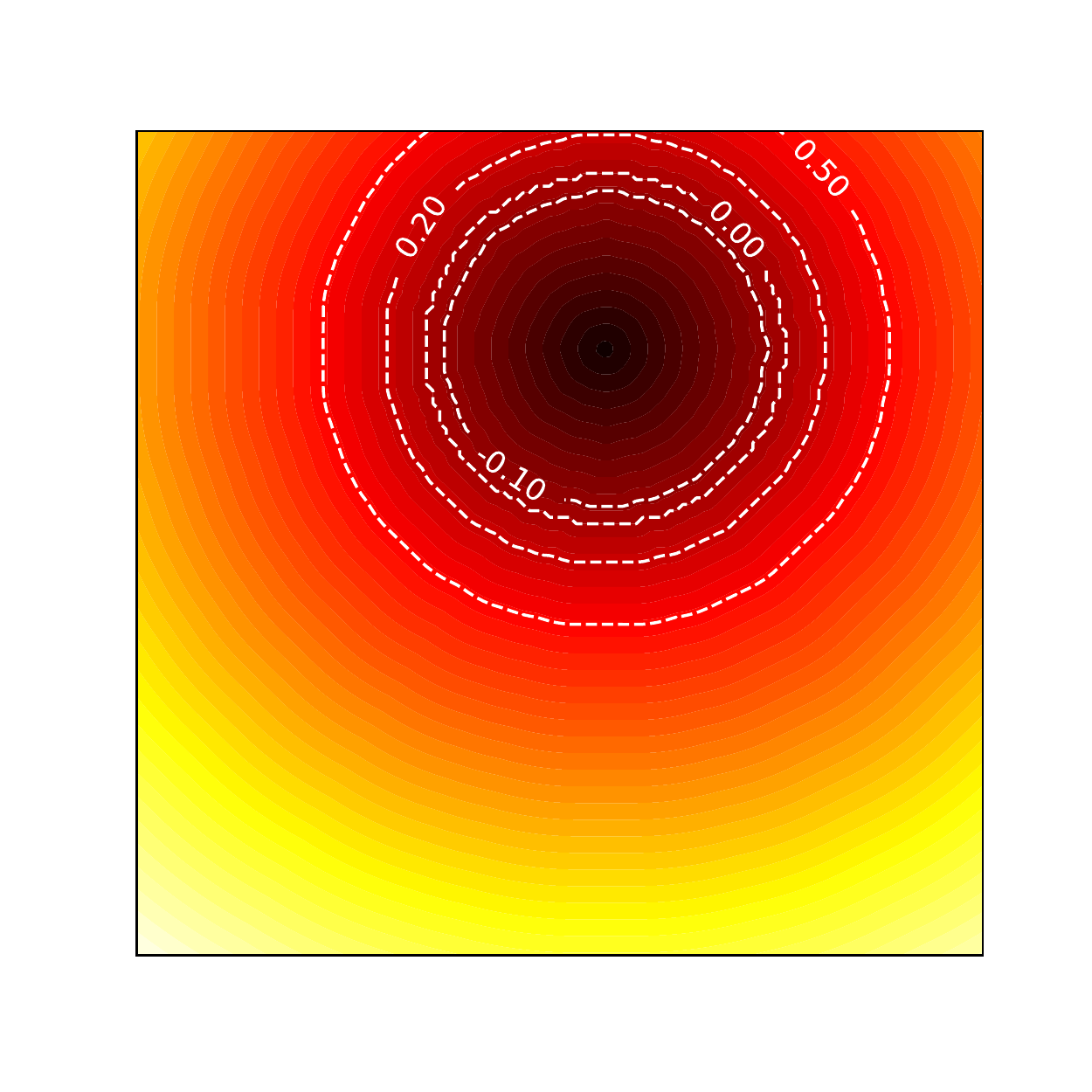}
	\hspace{0.1cm}
	\includegraphics[trim=0cm 1cm 1cm 1cm, clip=true, scale=0.20]{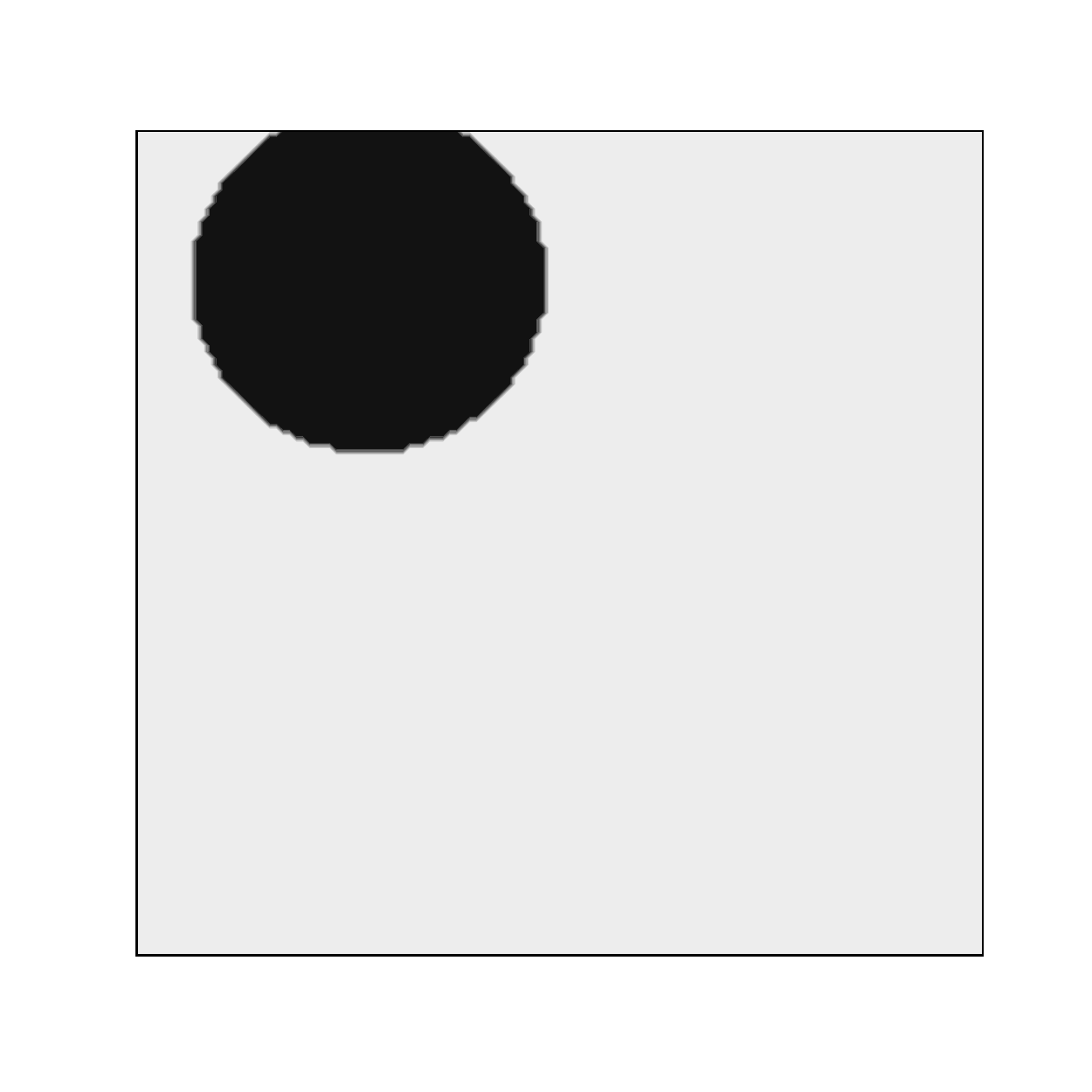}
	\hspace{-0.5cm}
	\includegraphics[trim=0cm 1cm 1cm 1cm, clip=true, scale=0.20]{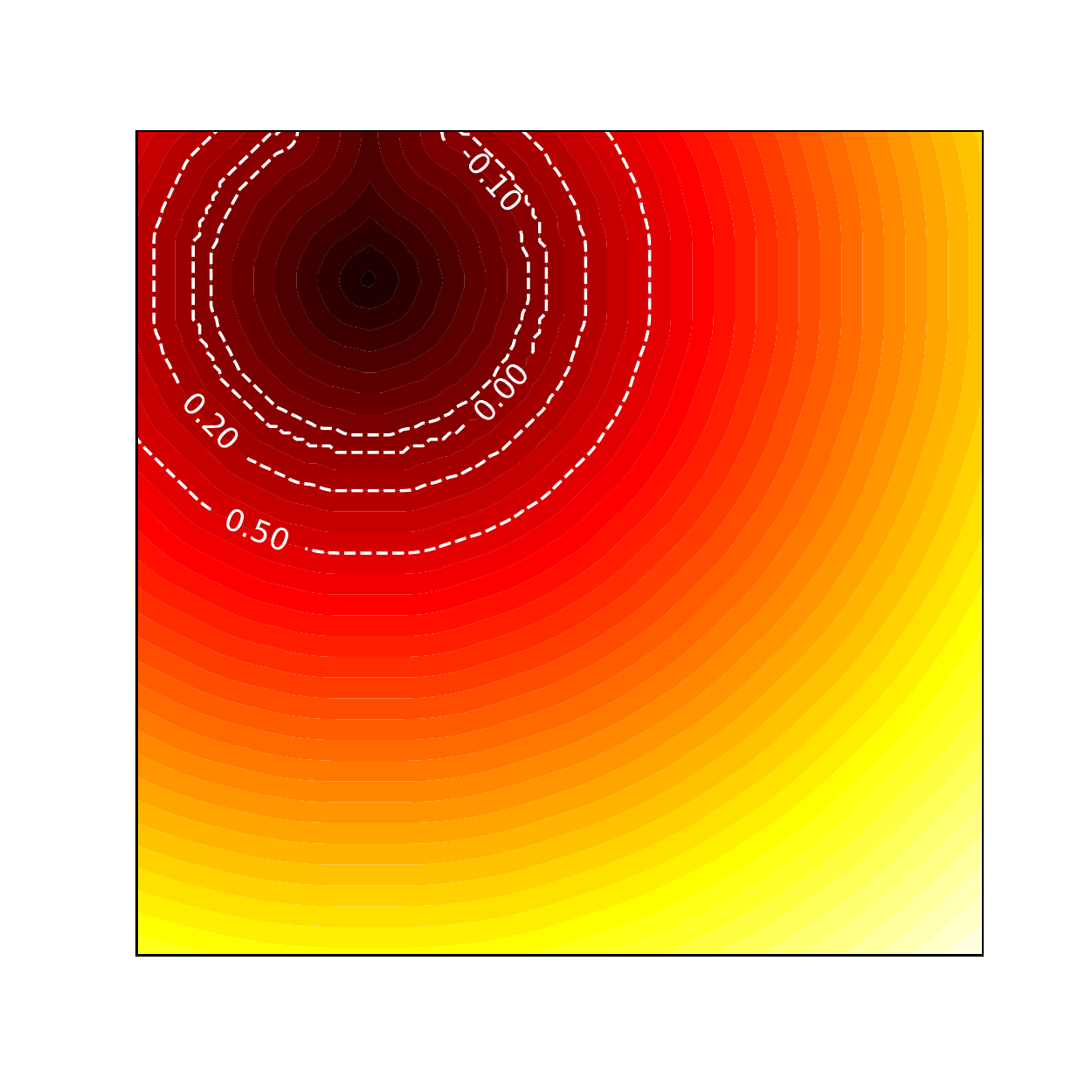}
	\hspace{0.1cm}
	\includegraphics[trim=0cm 1cm 1cm 1cm, clip=true, scale=0.20]{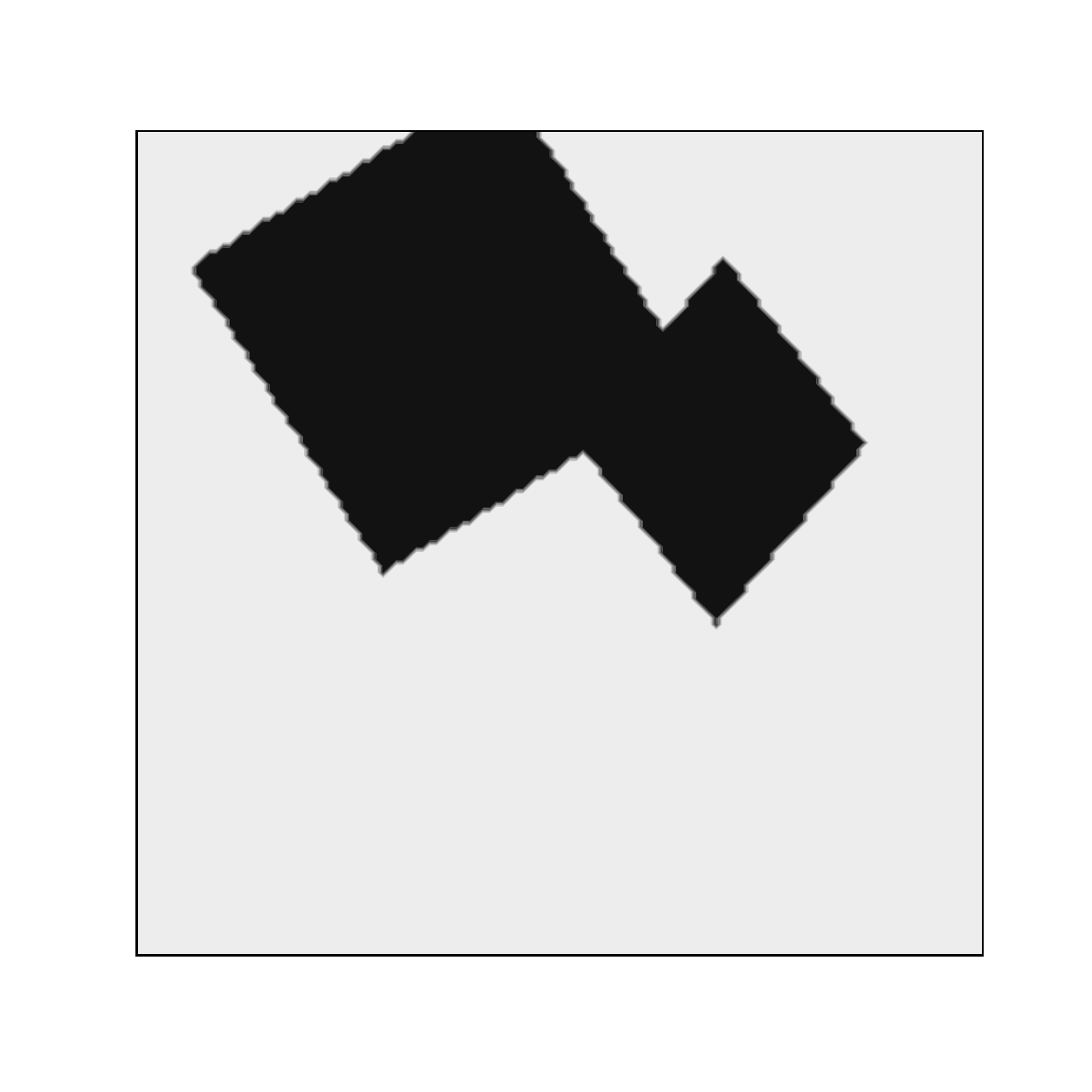}
	\hspace{-0.5cm}
	\includegraphics[trim=0cm 1cm 1cm 1cm, clip=true, scale=0.20]{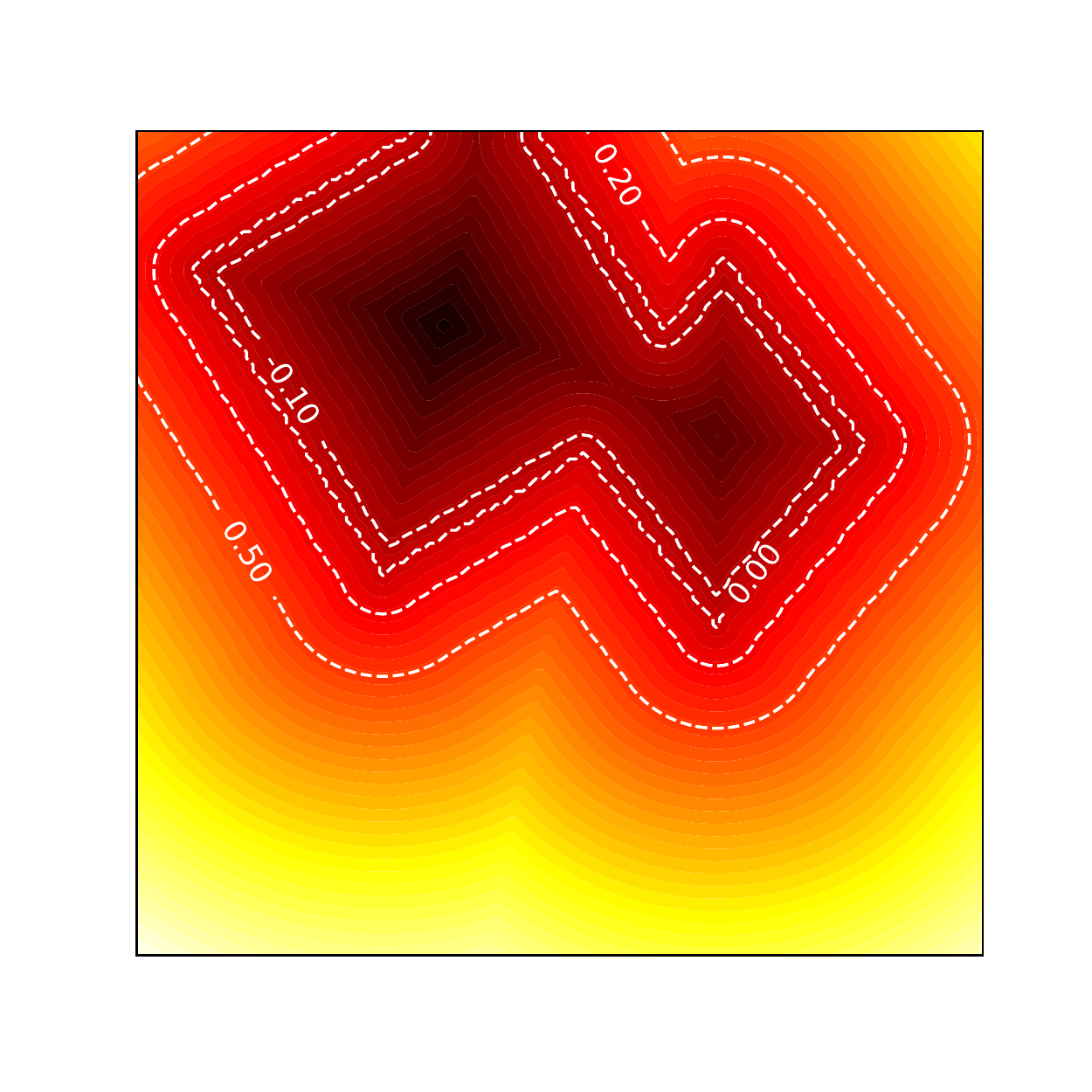}\\
	\includegraphics[trim=0cm 1cm 1cm 1cm, clip=true, scale=0.20]{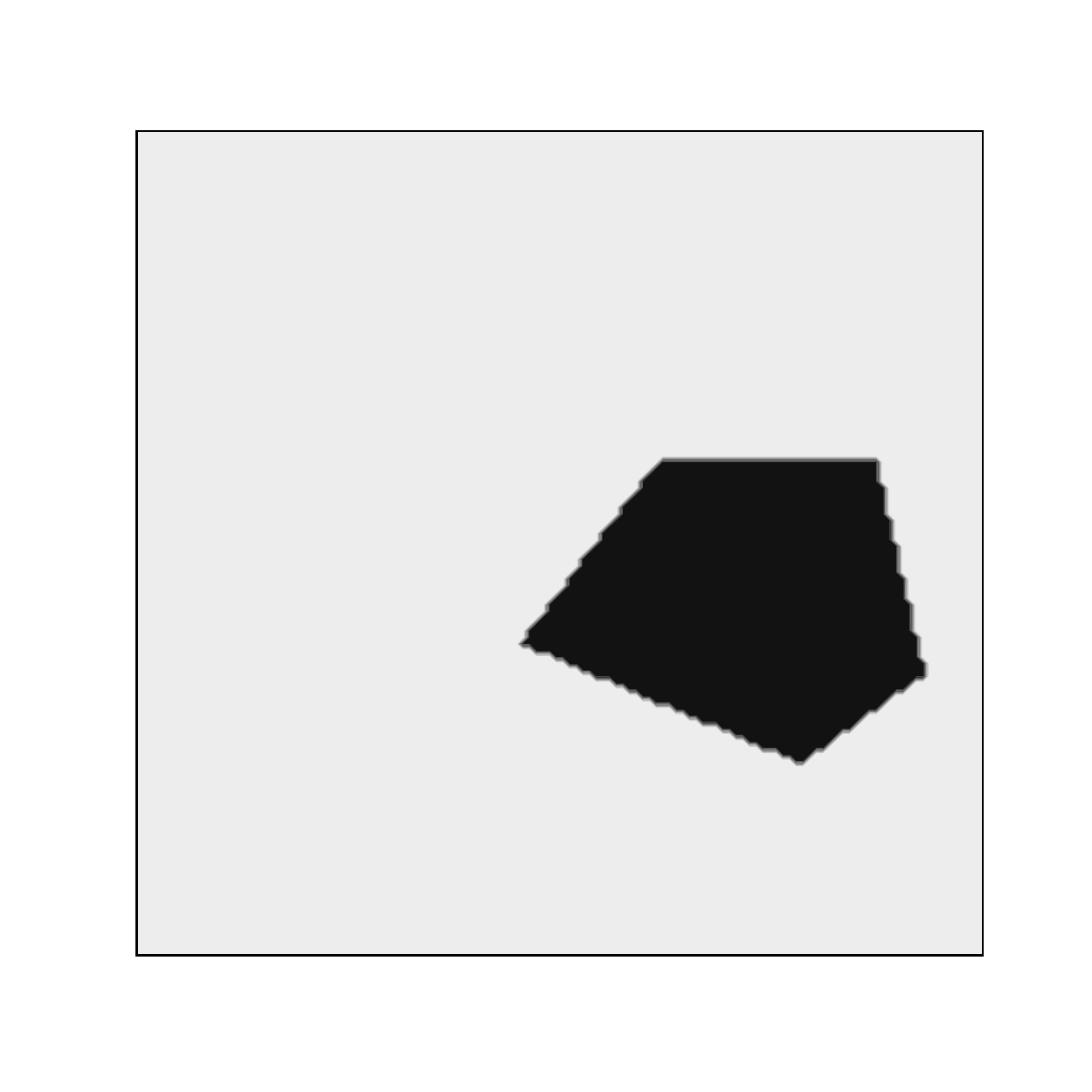}
	\hspace{-0.5cm}
	\includegraphics[trim=0cm 1cm 1cm 1cm, clip=true, scale=0.20]{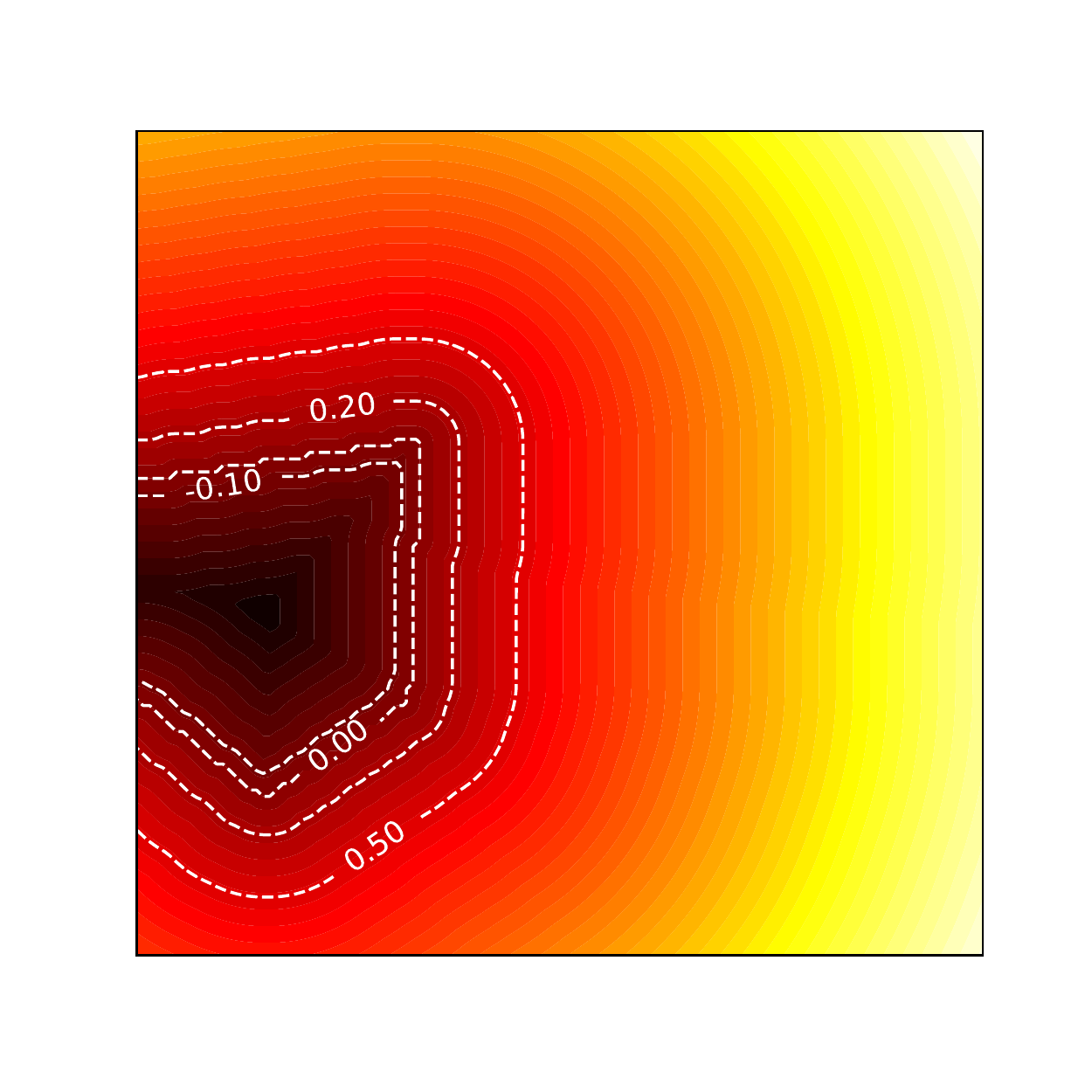}
	\hspace{0.1cm}
	\includegraphics[trim=0cm 1cm 1cm 1cm, clip=true, scale=0.20]{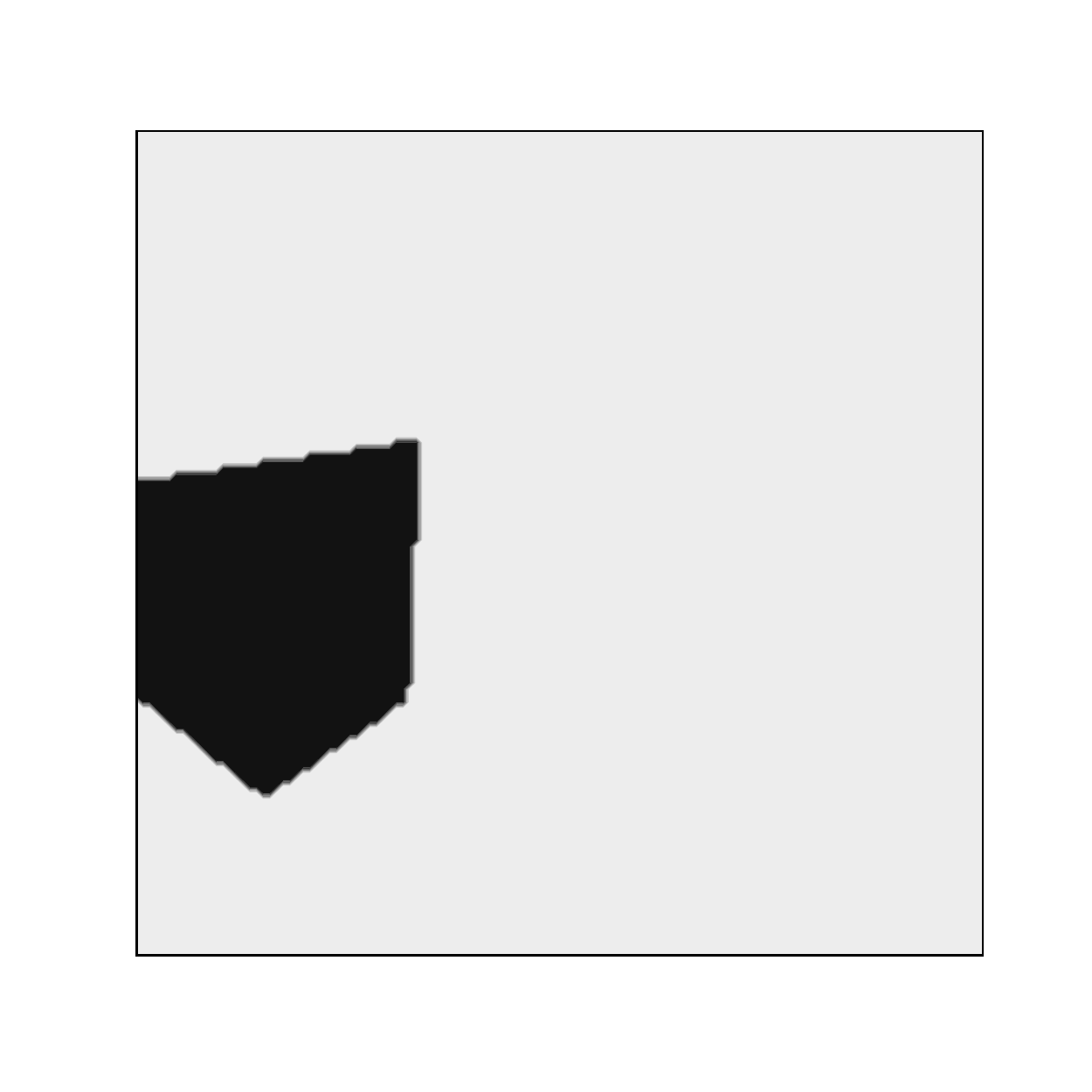}
	\hspace{-0.5cm}
	\includegraphics[trim=0cm 1cm 1cm 1cm, clip=true, scale=0.20]{figs/img_sdf_4}
	\hspace{0.1cm}
	\includegraphics[trim=0cm 1cm 1cm 1cm, clip=true, scale=0.20]{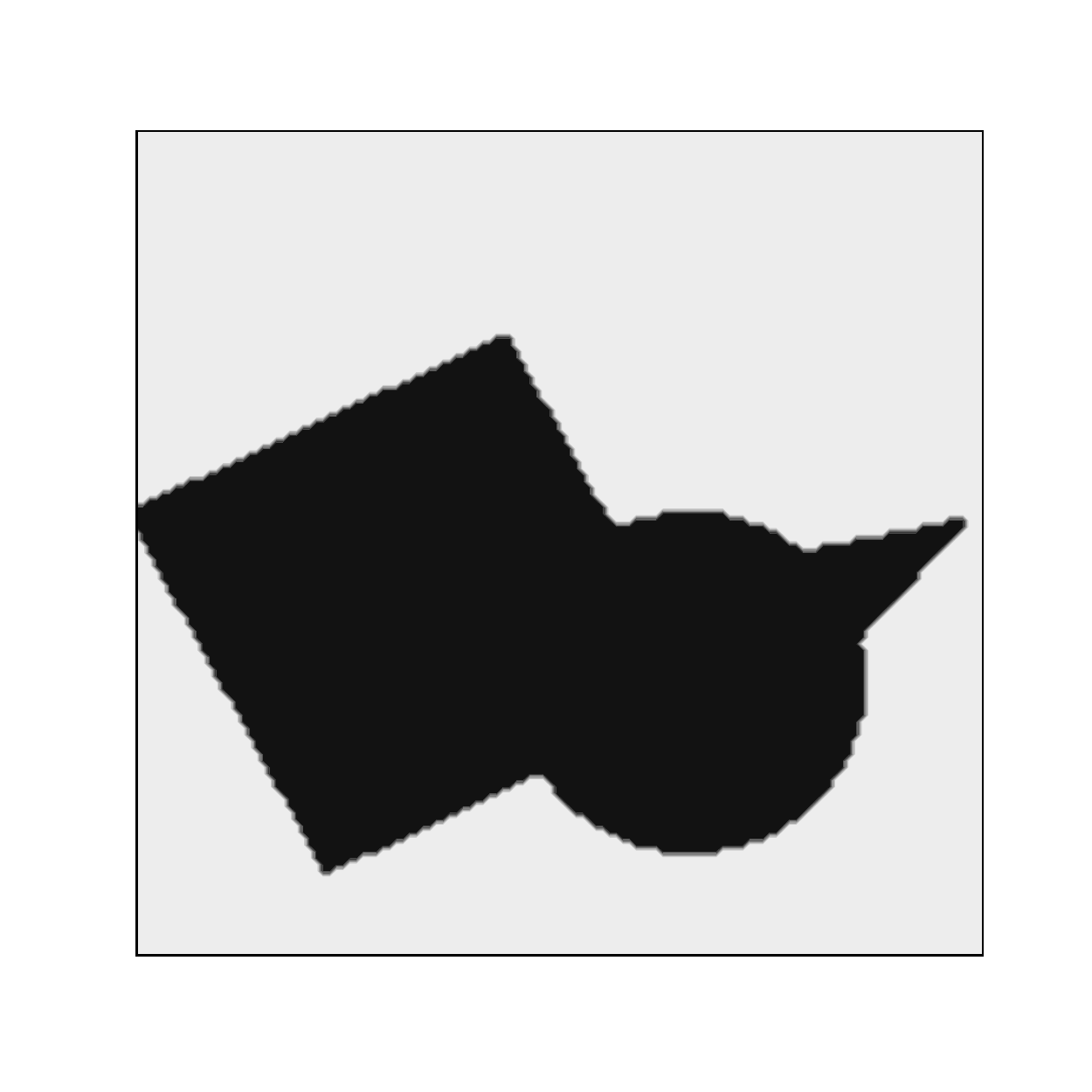}
	\hspace{-0.5cm}
	\includegraphics[trim=0cm 1cm 1cm 1cm, clip=true, scale=0.2]{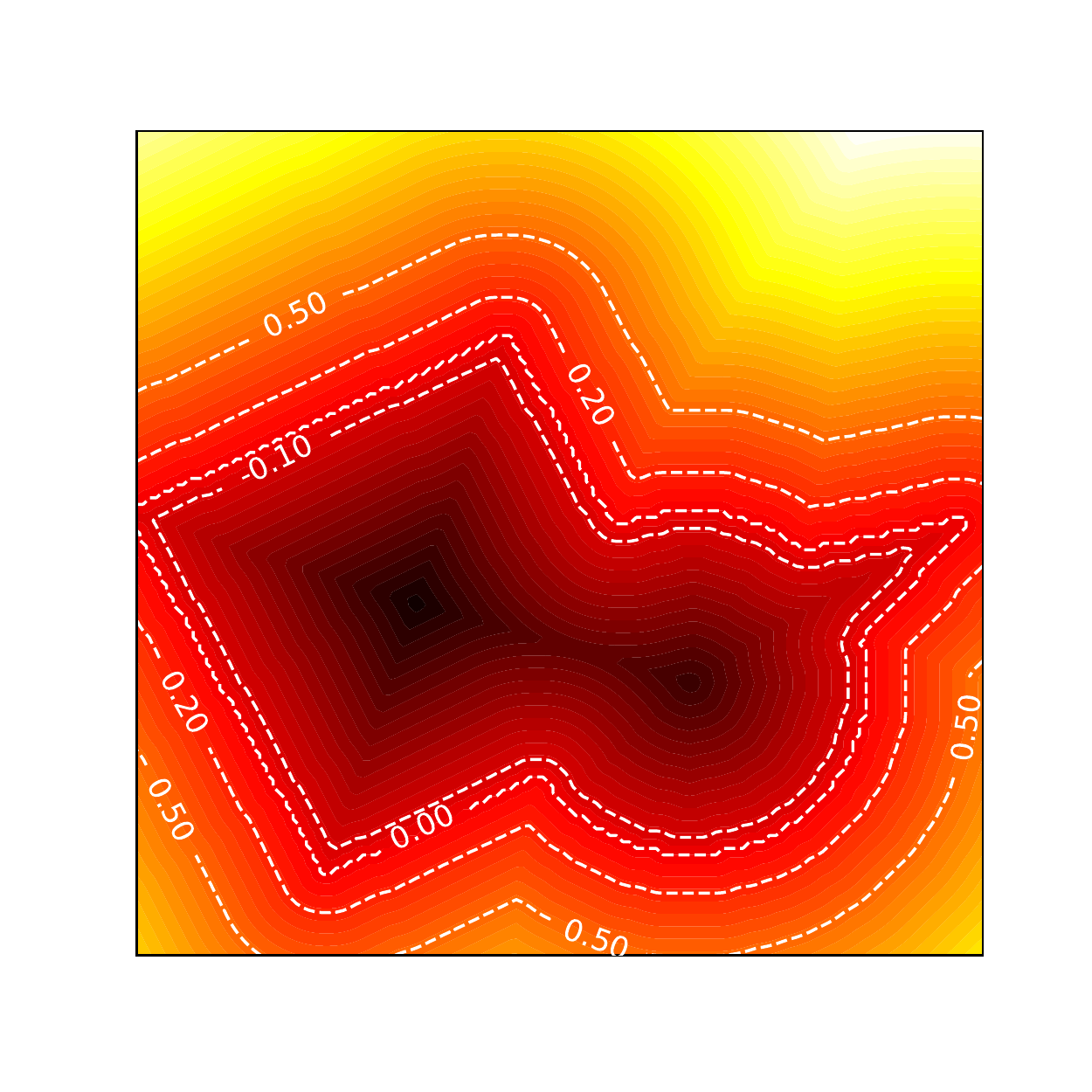}
\end{tabular}
\caption{Data generation process. left column) training data contains primitive shapes and the associated SDF; middle column) training dataset is augmented by random rotation, translation and scaling; right column) training dataset is augmented by random combination of 2 or 3 objects.\label{fig:datagen}}
\end{figure}


The processor and compressor networks are trained using data generated with primitive 2D shapes: circles, triangles, rectangles and polygons; see Fig.~\ref{fig:datagen}. The SDF is computed following two distance transformations, as described in \cite{sitzmann2020metasdf} \footnote{The SDF for the primitive shapes can be computed analytically; see for example \href{https://www.iquilezles.org/www/articles/distfunctions2d/distfunctions2d.htm}{here} . But as explained \href{https://www.iquilezles.org/www/articles/interiordistance/interiordistance.htm}{here}, there is no analytical way to combine analytical expressions of multiple SDFs to generate a new valid SDF. Therefore, we rely on numerical estimation of SDFs.}. To enrich the dataset, we have augmented geometries by random rotating, translating or scaling of the primitive shapes, as well as by combining two or three shapes together, see Fig.~\ref{fig:datagen}.
The input binary image and associated SDF are  $128\times 128$.

The processor and compressor networks are trained in a supervised learning
fashion (recall our evalutor network does not have any learning parameters). Following \cite{park2019deepsdf}, we use mean absolute error (MAE) as the loss function. The training is performed with ADAM optimizer \citep{kingma2014adam} with an initial learning rate of $5\times 10^{-4}$ and momentum parameters $(\beta_1, \beta_2) =(0.9, 0.999)$. The learning rate was reduced as training progressed. The batch size and number of epochs varied for each network. We found that training the networks separately will lead to more stable training. Therefore, we initially trained the processor, froze the weights, and then trained the compressor. 

\subsection{Results on complex geometries}\label{sec:exotic}
\begin{figure}
\centering
\begin{tabular}{cc}
    {\hspace{-1.5cm} binary image \hspace{2.5cm} ground truth \hspace{3cm} processor}\\
	\includegraphics[trim=0cm 1cm 1cm 1cm, clip=true, scale=0.4]{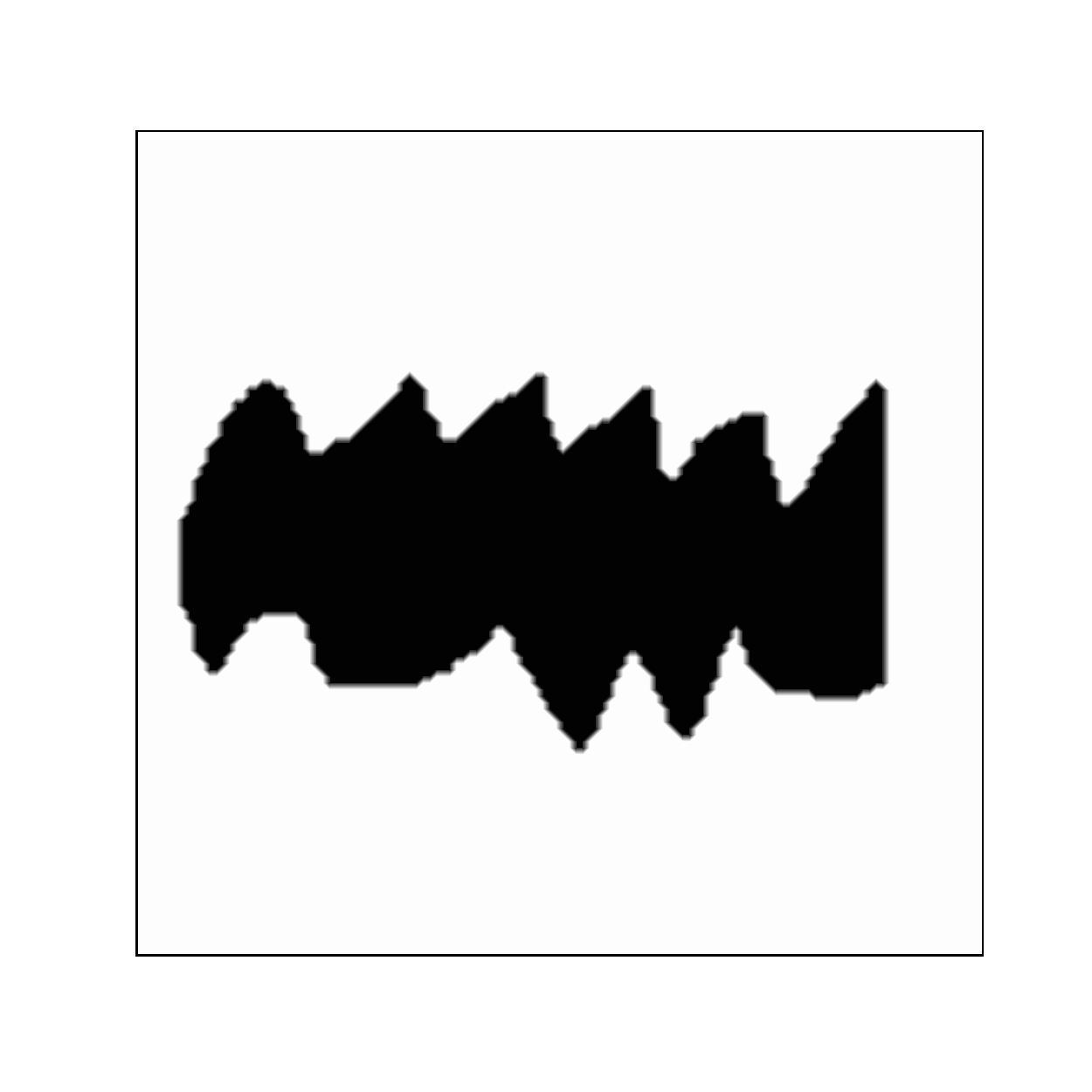}
	\includegraphics[trim=0cm 1cm 1cm 1cm, clip=true, scale=0.4]{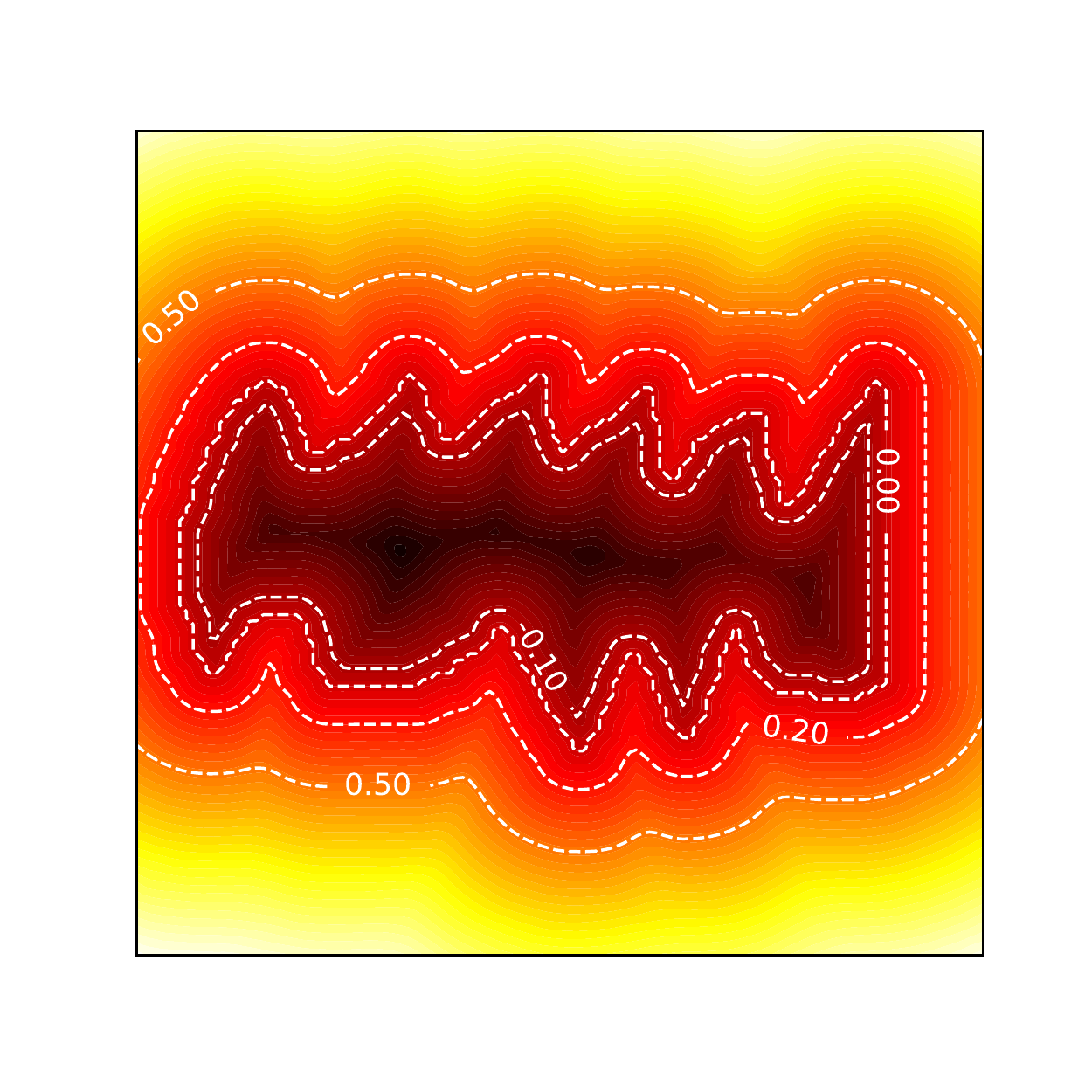}
	\includegraphics[trim=0cm 1cm 1cm 1cm, clip=true, scale=0.4]{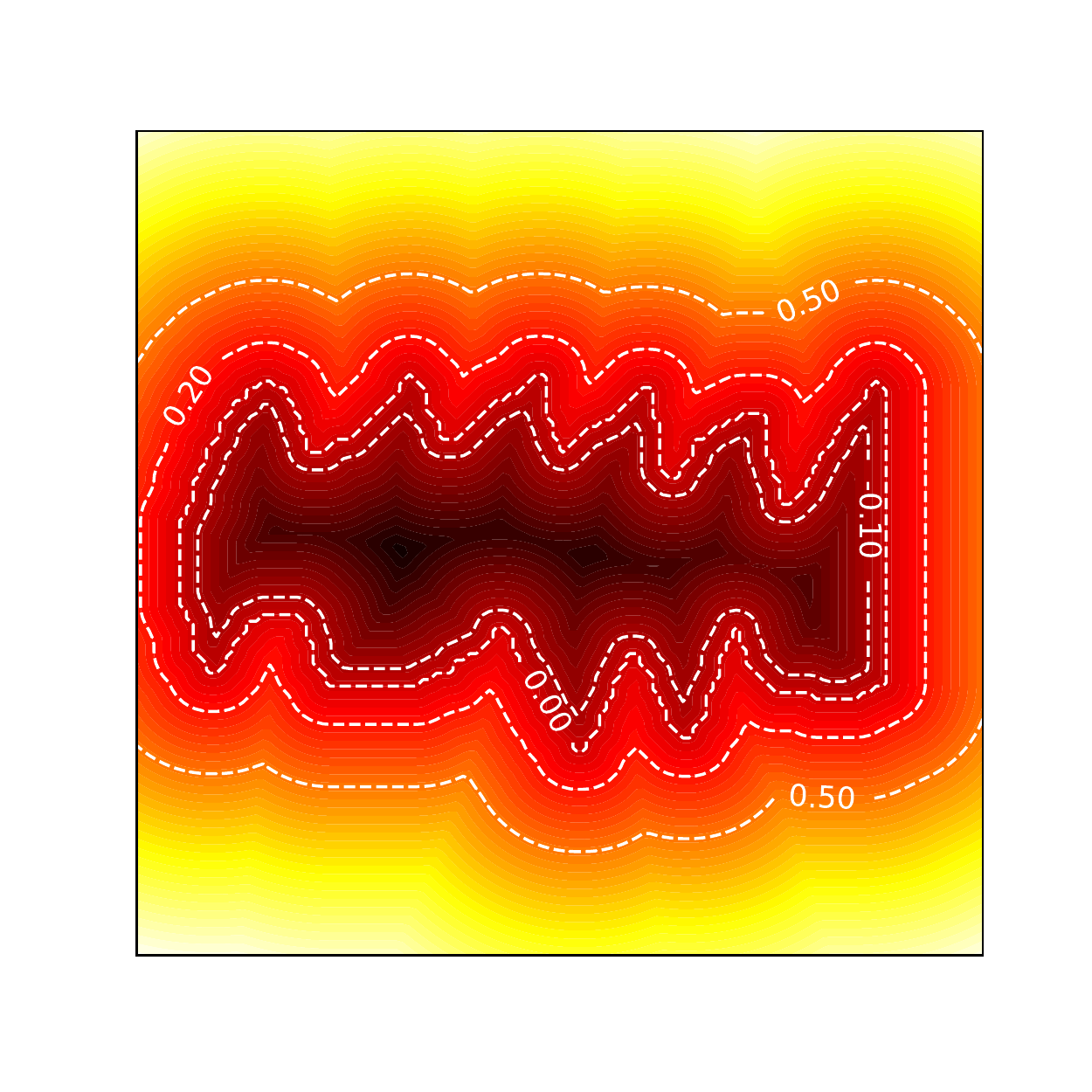}\\
	\includegraphics[trim=0cm 1cm 1cm 1cm, clip=true, scale=0.4]{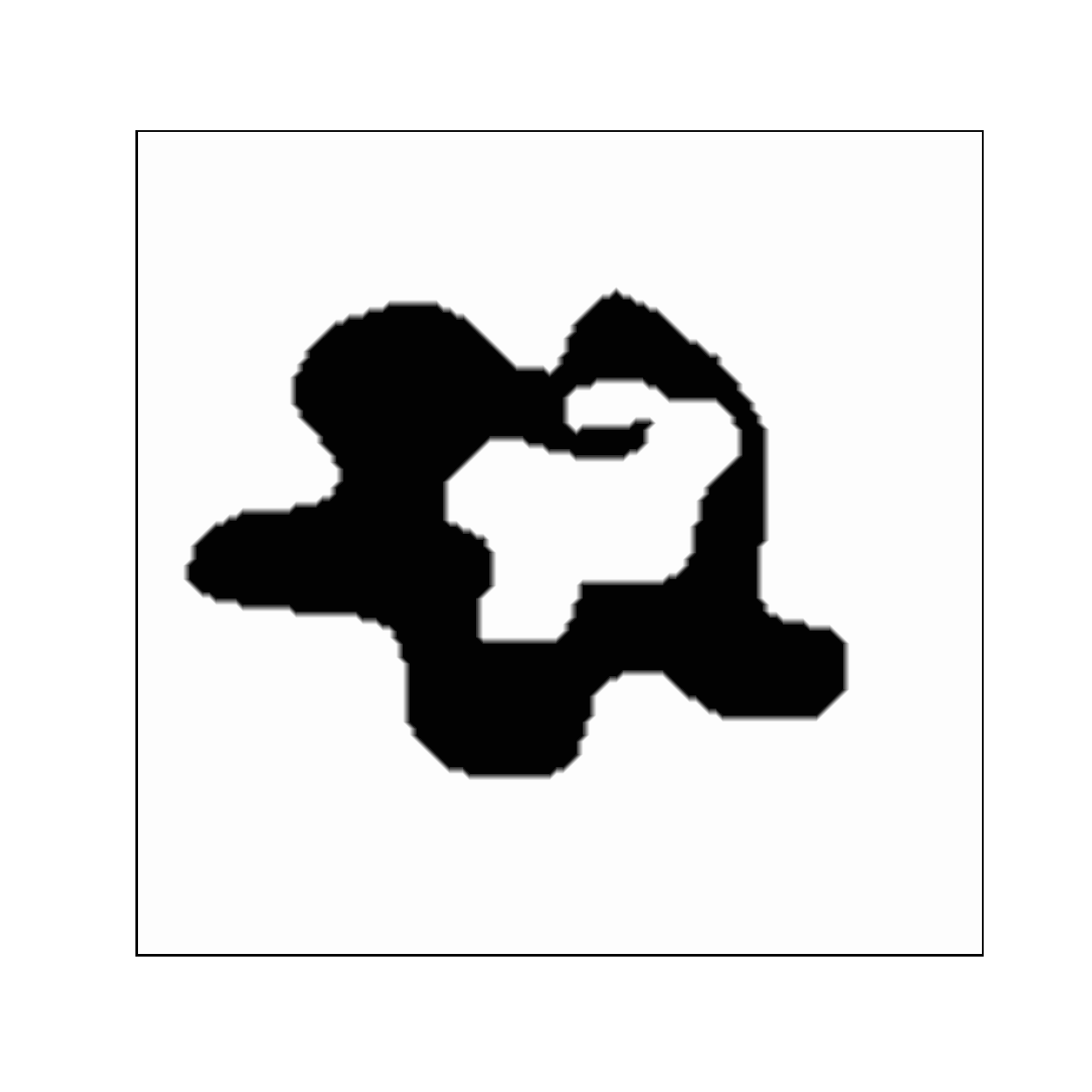}
	\includegraphics[trim=0cm 1cm 1cm 1cm, clip=true, scale=0.4]{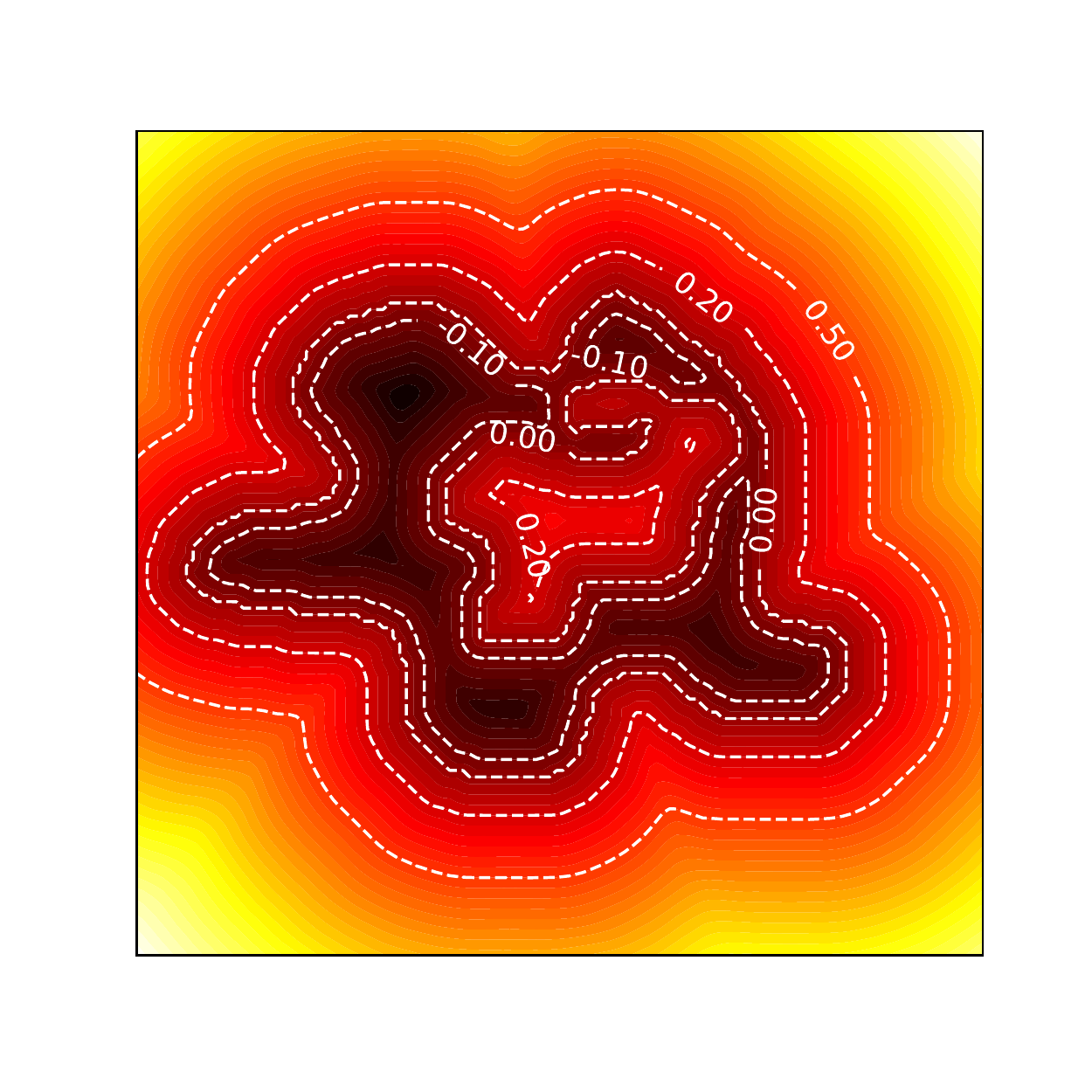}
	\includegraphics[trim=0cm 1cm 1cm 1cm, clip=true, scale=0.4]{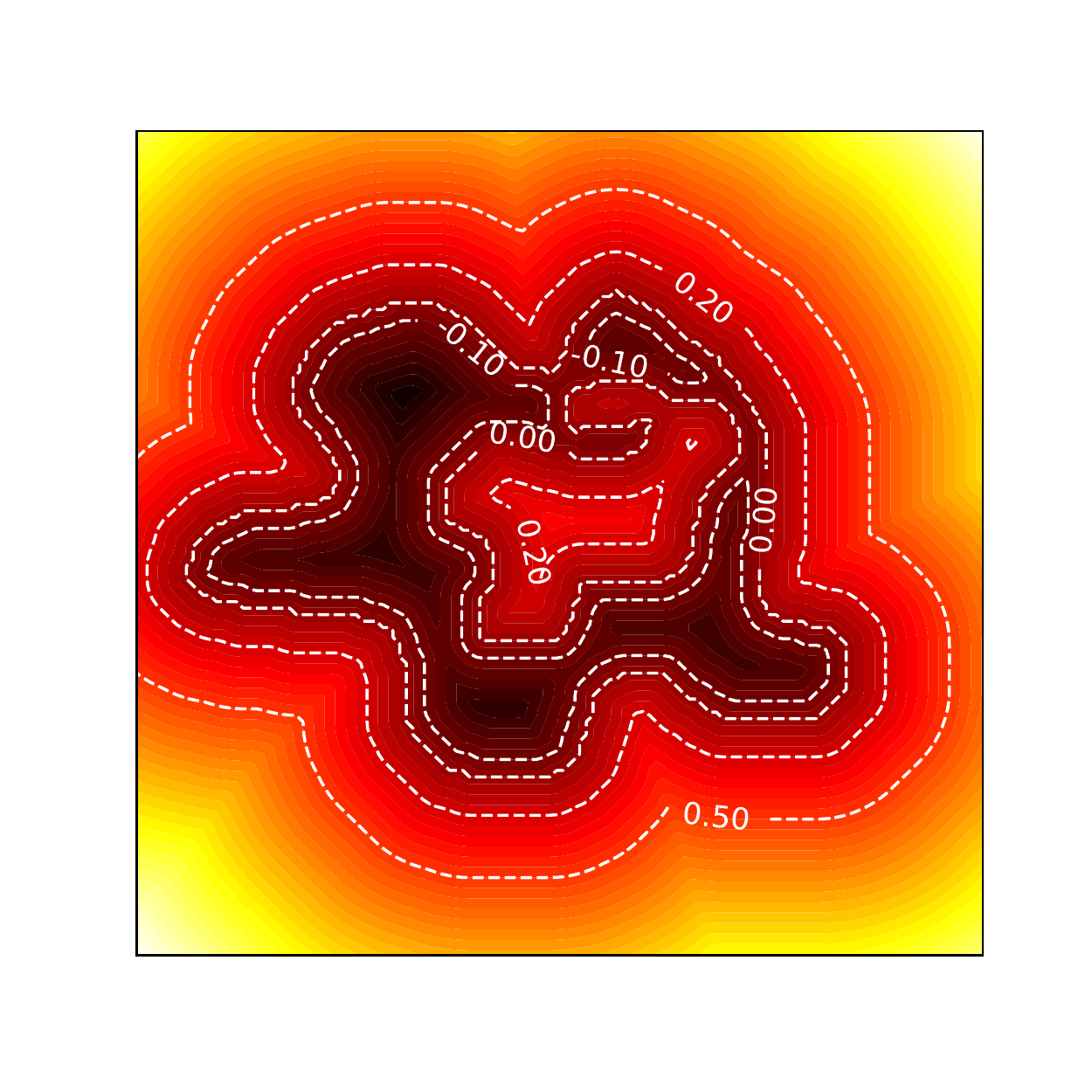}\\
	\includegraphics[trim=0cm 1cm 1cm 1cm, clip=true, scale=0.4]{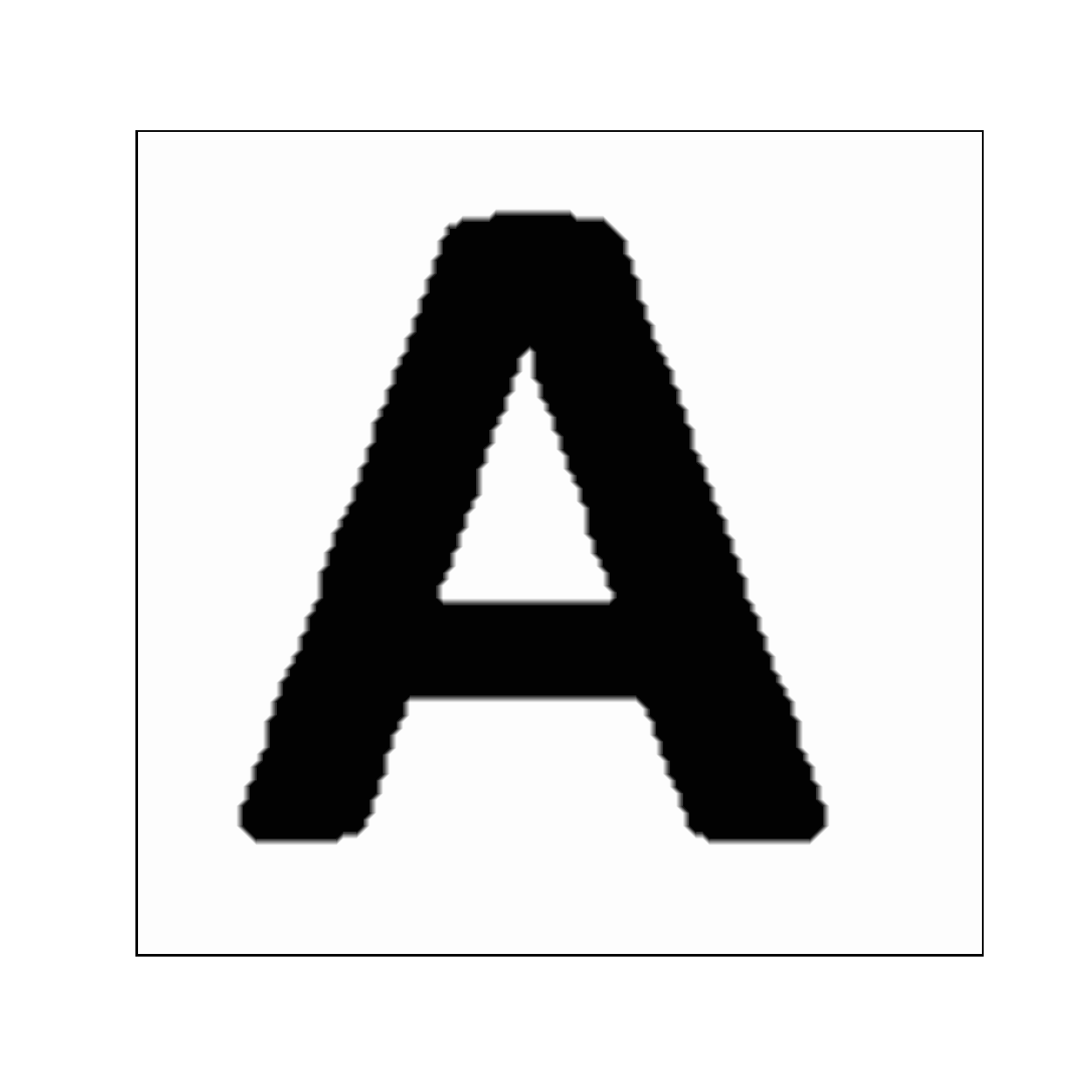}
	\includegraphics[trim=0cm 1cm 1cm 1cm, clip=true, scale=0.4]{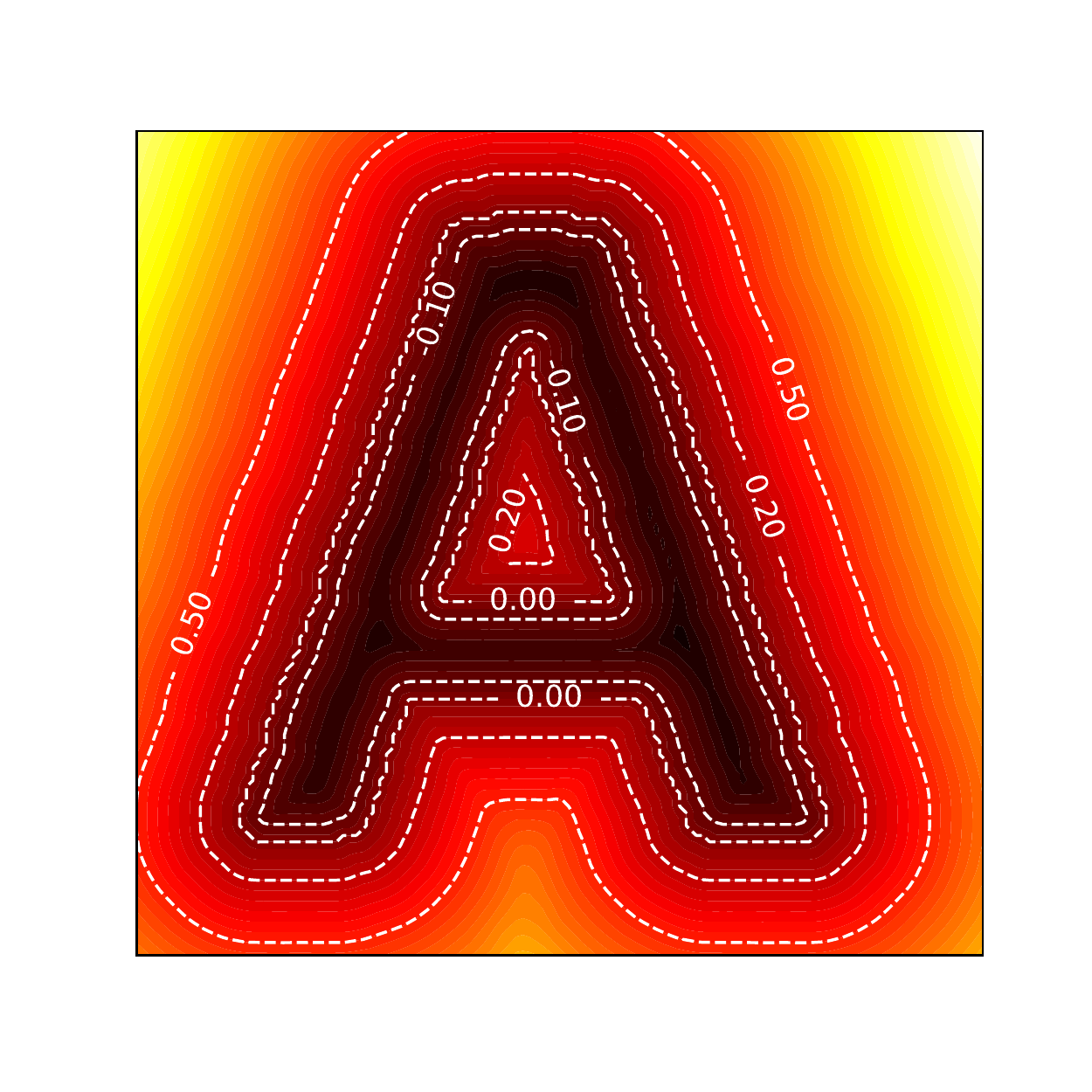}
	\includegraphics[trim=0cm 1cm 1cm 1cm, clip=true, scale=0.4]{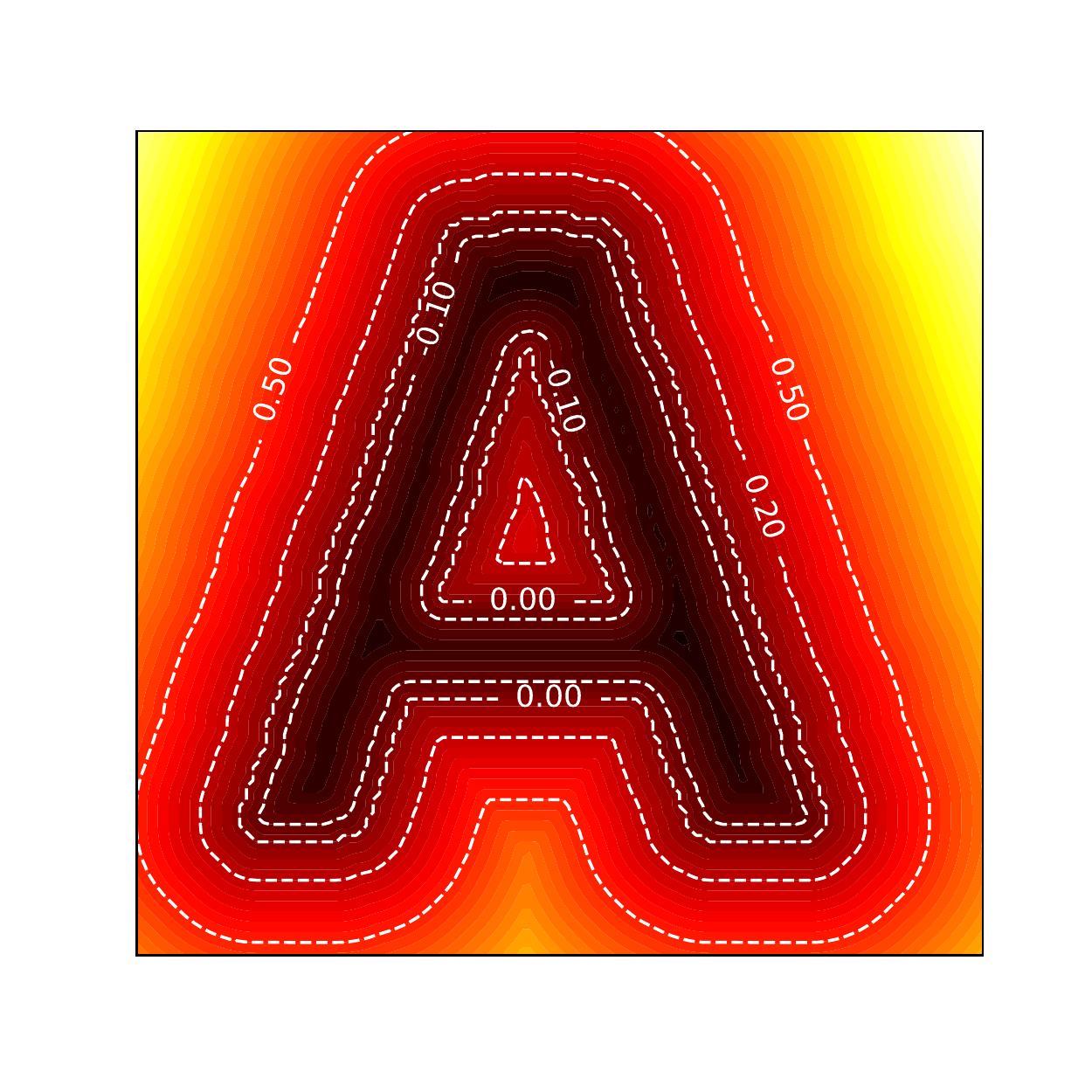}\\
\end{tabular}
\caption{Performance of the processor on more complex geometries. The compressor produces almost exact same results.}
\label{fig:results_exotic}
\end{figure}

Although the training set contains only simple shapes, the processor and compressor are able to accurately predict the SDF of complex geometries. Example of such complex shapes are presented in Fig. \ref{fig:results_exotic}

\end{document}